%% file: main.tex
\documentclass[10pt,twocolumn,letterpaper]{article}
\usepackage[pagenumbers]{iccv}               

\definecolor{iccvblue}{rgb}{0.21,0.49,0.74}
\usepackage[pagebackref,breaklinks,colorlinks,allcolors=iccvblue]{hyperref}

\usepackage{multirow}
\usepackage{colortbl}%
\newcommand{\myrowcolour}{\rowcolor[gray]{0.925}}

\definecolor{LightGray}{gray}{0.925}
\newcommand{\secondnum}[1]{\textcolor{blue}{\underline{#1}}}
\newcommand{\bestnum}[1]{\textcolor{red}{#1}}
\usepackage{makecell}
\usepackage{bbding}
\usepackage{algorithmic}
\usepackage{algorithm}

\newcommand\ours{FrDiff\xspace}

\title{Frequency Domain-Based Diffusion Model for Unpaired Image Dehazing}
\author{Chengxu Liu\textsuperscript{1}\quad\quad Lu Qi\textsuperscript{2,3}\quad\quad Jinshan Pan\textsuperscript{4}\quad\quad Xueming Qian\textsuperscript{1}\quad\quad Ming-Hsuan Yang\textsuperscript{5}\\
\textsuperscript{1}Xi’an Jiaotong University\quad 
\textsuperscript{2}Wuhan University\quad
\textsuperscript{3}Insta360 \\
\textsuperscript{4}Nanjing University of Science and Technology\quad
\textsuperscript{5}University of California, Merced 
\\
}

\begin{document}
\maketitle
\begin{abstract}
Unpaired image dehazing has attracted increasing attention due to its flexible data requirements during model training.
Dominant methods based on contrastive learning not only introduce haze-unrelated content information, but also ignore haze-specific properties in the frequency domain (\ie,~haze-related degradation is mainly manifested in the amplitude spectrum).
To address these issues, we propose a novel frequency domain-based diffusion model, named \ours, for fully exploiting the beneficial knowledge in unpaired clear data.
In particular, inspired by the strong generative ability shown by Diffusion Models (DMs), we tackle the dehazing task from the perspective of frequency domain reconstruction and perform the DMs to yield the amplitude spectrum consistent with the distribution of clear images.
To implement it, we propose an Amplitude Residual Encoder (ARE) to extract the amplitude residuals, which effectively compensates for the amplitude gap from the hazy to clear domains, as well as provide supervision for the DMs training.
In addition, we propose a Phase Correction Module (PCM) to eliminate artifacts by further refining the phase spectrum during dehazing with a simple attention mechanism.
Experimental results demonstrate that our \ours outperforms other state-of-the-art methods on both synthetic and real-world datasets.
\end{abstract}

\section{Introduction}
\label{sec:intro}

\begin{figure}[t]
  \centering
    \includegraphics[width=1.0\linewidth,page=1]{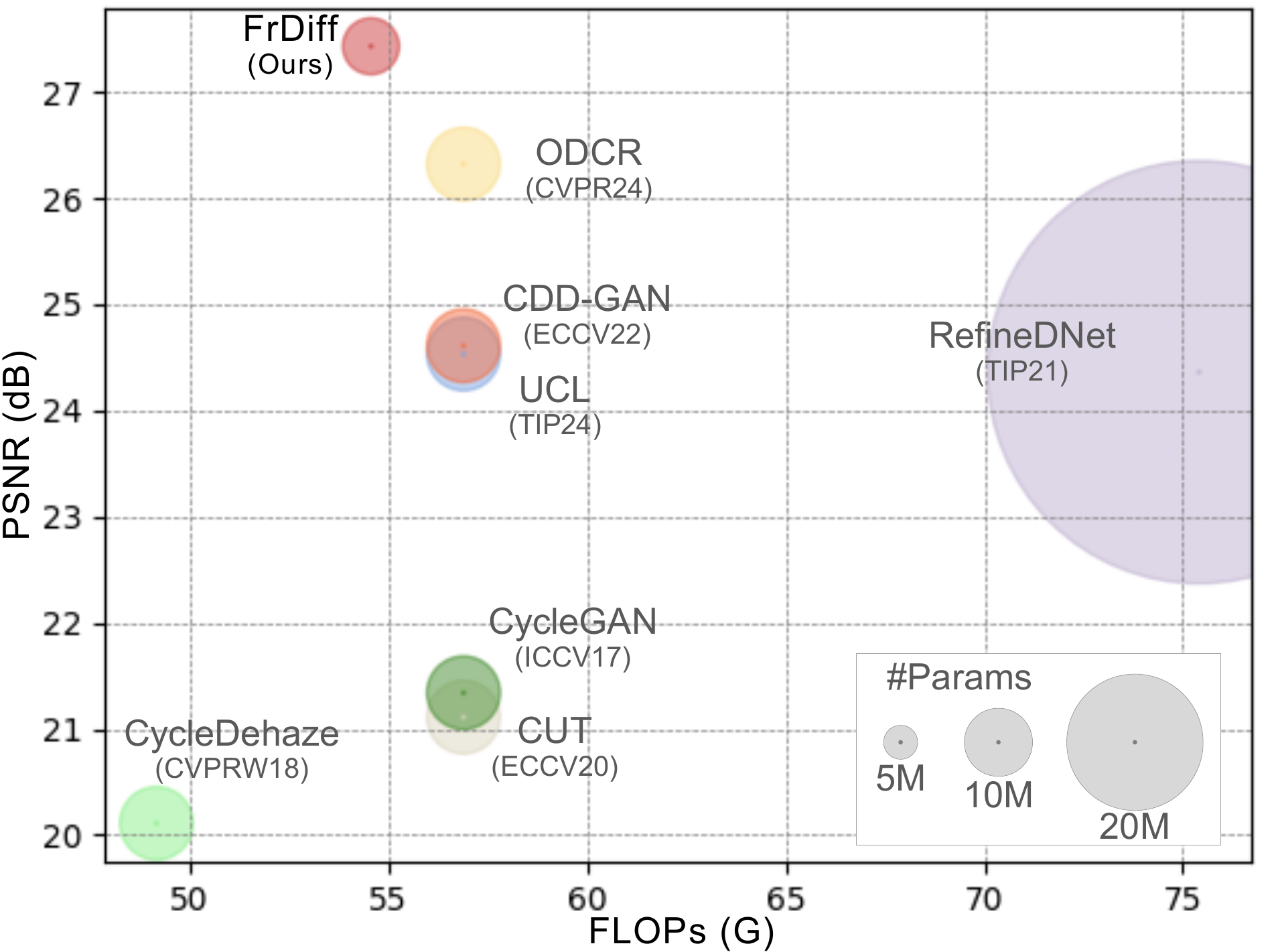}
    \vspace{-6mm}
   \caption{Comparison of performance, FLOPs, and parameters with state-of-the-art methods on the SOTS-Indoor dataset~\cite{li2018benchmarking}. Notably, \ours is the first work to apply the diffusion model to the unpaired image dehazing task. Better viewed in color.}
   \label{fig:performance}
\end{figure}

\begin{figure*}[t]
  \centering
    \includegraphics[width=1.0\linewidth,page=2]{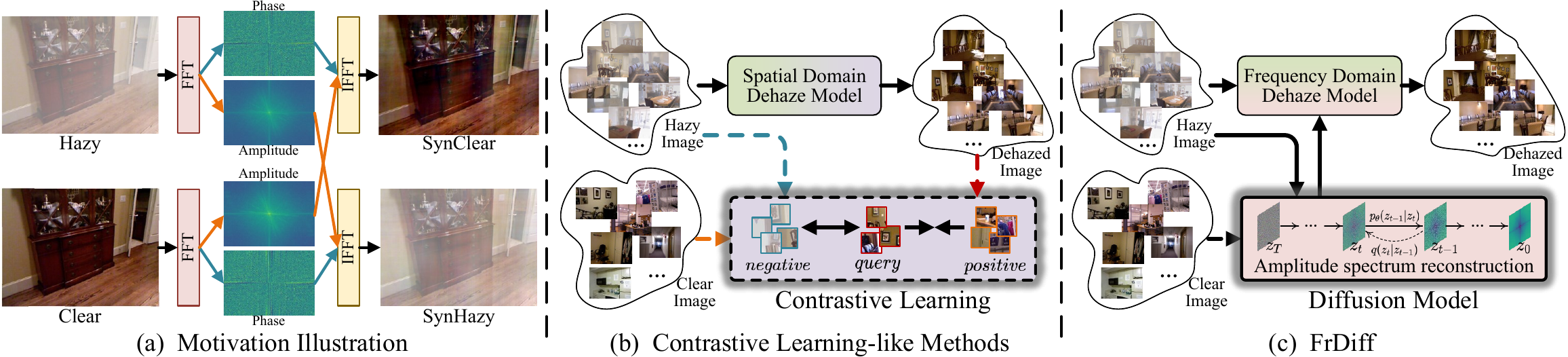}
    \vspace{-7mm}
   \caption{(a) Illustration of the property of haze degradation in the frequency domain.
   Haze degradation can be transferred with the exchange of the amplitude spectrum. (b) Contrastive learning-like methods construct positive and negative sample pairs to maximize the mutual information between clean and hazy images. (c) Our \ours learns the amplitude spectrum of clear images from unpaired data during training and reconstructs them during inference.}
   \vspace{-2mm}
   \label{fig:teaser}
\end{figure*}

Haze is an often-occurring atmospheric phenomenon caused by the scattering effect of aerosol particles, such as dust, smoke, and others. It usually leads to scene radiative attenuation, producing a severe degradation of visual content~\cite{he2010single}. 
Image dehazing can be applied to improve the performance of subsequent high-level vision tasks~\cite{li2023detection,sakaridis2018model}.

To remove the haze, conventional methods manually designed image priors based on the atmospheric scattering model~\cite{narasimhan2002vision} or observations~\cite{he2010single,fattal2014dehazing,zhu2015fast,berman2018single,berman2016non}. However, these hand-crafted priors struggled to accurately model the intrinsic features of haze and often lacked robustness and reliability. 
Recent deep learning-based methods focus mainly on developing supervised models using synthetic paired hazy-clear data~\cite{ren2016single,zhang2018densely,qin2020ffa,dong2020multi,liu2019griddehazenet,yu2022frequency}.  
Among them, the finding that the illuminance contrast (\eg,~haze) is manifested by the amplitude spectrum~\cite{skarbnik2009importance,oppenheim1981importance} shows great potential in haze removal through the frequency domain~\cite{yu2022frequency,li2023embedding,qiao2023learning}
(illustrated by Fig.~\ref{fig:teaser}(a), please refer to the supplementary for detailed statistical analyses).
Nevertheless, the large domain gap between synthetic haze and realistic haze limits its generalization ability in real-world scenarios~\cite{yang2022self}. Moreover, acquiring paired hazy-clear data is prohibitively expensive and time-consuming.

To tackle these problems, an increasing number of unpaired image dehazing methods have emerged.
Among these, CycleGAN-like methods~\cite{qiao2023learning,liu2020end,engin2018cycle,yang2022self} attempt to establish a cyclic transformation between the hazy and clear domains. 
Unfortunately, the diversity in haze intensity often leads to asymmetric domain knowledge. 
Such a bijective relationship between the two domains can lead to sub-optimal performance~\cite{park2020contrastive}. 
Additionally, some methods inspired by contrastive learning~\cite{park2020contrastive,chen2022unpaired,wang2024ucl,wang2024odcr} focus on maximizing the mutual information between the dehazed output and the hazy input, as illustrated by~\cref{fig:teaser}(b).
However, such practice also introduces haze-unrelated content information and tends to overlook the frequency domain properties of haze, leading to unsatisfactory results.

Recently, diffusion models (DMs) have demonstrated impressive performance in image restoration~\cite{xia2023diffir,chen2024hierarchical,kawar2022denoising,lugmayr2022repaint,li2024rethinking,ye2024learning}, which has prompted us to investigate their effectiveness in haze removal. However, applying these models to unpaired image dehazing presents some challenges:
1) DMs are sensitive to luminance contrast. Methods that directly utilize DMs to generate clear images~\cite{kawar2022denoising,lugmayr2022repaint} often lead to unpredictable artifacts and color aberrations, while also incurring significant computational costs~\cite{yu2023high,ye2024learning}.
2) Other methods~\cite{xia2023diffir,chen2024hierarchical,li2024rethinking} that use DMs to generate latent representations can only learn from high-quality paired data, which restricts their application in unpaired settings.
As a result, exploring effective ways to leverage DMs for unpaired image restoration remains a significant challenge.

In this paper, we concentrate on the distinctive frequency domain characteristics of haze and propose a novel diffusion model for unpaired image dehazing (\ours).
The core idea behind \ours is to employ DM learning to reconstruct the amplitude of clear images from unpaired data, as illustrated in~\cref{fig:teaser}(c).
Specifically, we divide the training process of \ours into two steps following the previous practice~\cite{xia2023diffir,chen2024hierarchical} as illustrated in~\cref{fig:training}.
In the first step, we introduce the Amplitude Residual Encoder (ARE), which extracts amplitude residuals to address the gaps in amplitude distribution between hazy and clear images without interfering with haze-unrelated features.
In addition, we incorporate a Phase Correction Module (PCM) within the dehazing network to enhance the phase spectrum further. 
We optimize the dehazing network to effectively utilize the amplitude residuals for haze removal.
In the second step, we utilize the amplitude residuals obtained from the ARE as ground truth and train DM to reconstruct these residuals from Gaussian noise for haze removal.
During inference, the DM reconstructs the amplitude residuals using Gaussian noise as input and the amplitude spectrum of the hazy image as the condition. Compared to the task of reconstructing images that include illumination and texture structures, reconstructing amplitude residuals, which contain only illumination information, requires fewer iterations~\cite{oppenheim1981importance,skarbnik2009importance}.

Our contributions are summarized as follows:
\begin{itemize}
    \item We propose a novel frequency domain-based diffusion model for unpaired image dehazing (\ours), which is the first study to integrate a DM into unpaired restoration tasks. \ours enhances haze removal capabilities by learning amplitude reconstruction from unpaired data, providing inspiration for other unpaired restoration tasks.
    \item We propose an amplitude residual encoder (ARE) that generates amplitude residuals to bridge the gaps between hazy and clear domains without adding extra parameters, providing supervision for DM training.  
    \item We propose a phase correction module (PCM), designed to eliminate unwanted artifacts by refining the phase spectrum using an attention mechanism.
    \item Extensive experiments show that \ours outperforms existing SOTA methods with fewer parameters and FLOPs.
\end{itemize}

\section{Related Work}
\label{sec:relat}
\subsection{Single Image Dehazing}
\noindent\textbf{Prior-based Methods.} Early attempts aim to explore hand-crafted priors for haze removal by observing and analyzing the properties of hazy images, such as dark channel prior (DCP)~\cite{he2010single}, color attenuation prior (CAP)~\cite{zhu2015fast}, color-lines~\cite{fattal2014dehazing}, and haze-lines~\cite{berman2018single}. 
However, these approaches struggle to model varying degrees of haze simultaneously.

\vspace{1mm}
\noindent\textbf{Supervised learning Methods.} 
Based on paired data, these methods use DNNs to estimate atmospheric light and transmission maps~\cite{cai2016dehazenet,li2017aod,ren2016single,liu2019learning,zhang2018densely,wu2023ridcp} or directly learn hazy-to-clear translation~\cite{qin2020ffa,deng2020hardgan,dong2020multi,liu2019griddehazenet,ren2018gated,wu2021contrastive,zheng2023curricular,song2023vision,shao2020domain}. Typically, DehazeNet~\cite{cai2016dehazenet} and MSCNN~\cite{ren2016single} propose to estimate the transmission maps through an end-to-end model.
Then, GFN~\cite{ren2018gated}, MSDBN~\cite{dong2020multi}, and FFANet~\cite{qin2020ffa} develop a gated fusion network, dense feature fusion module, and feature fusion network to recover haze-free images, respectively. 
Recently, FSDGN~\cite{yu2022frequency} reveals the properties of haze in the frequency domain and provides inspiration for haze removal.
Although these techniques show superior performance, they tend to overfit the haze of synthetic data, resulting in sub-optimal performance in real-world scenarios.

\begin{figure}[t]
  \centering
\includegraphics[width=1.0\linewidth,page=4]{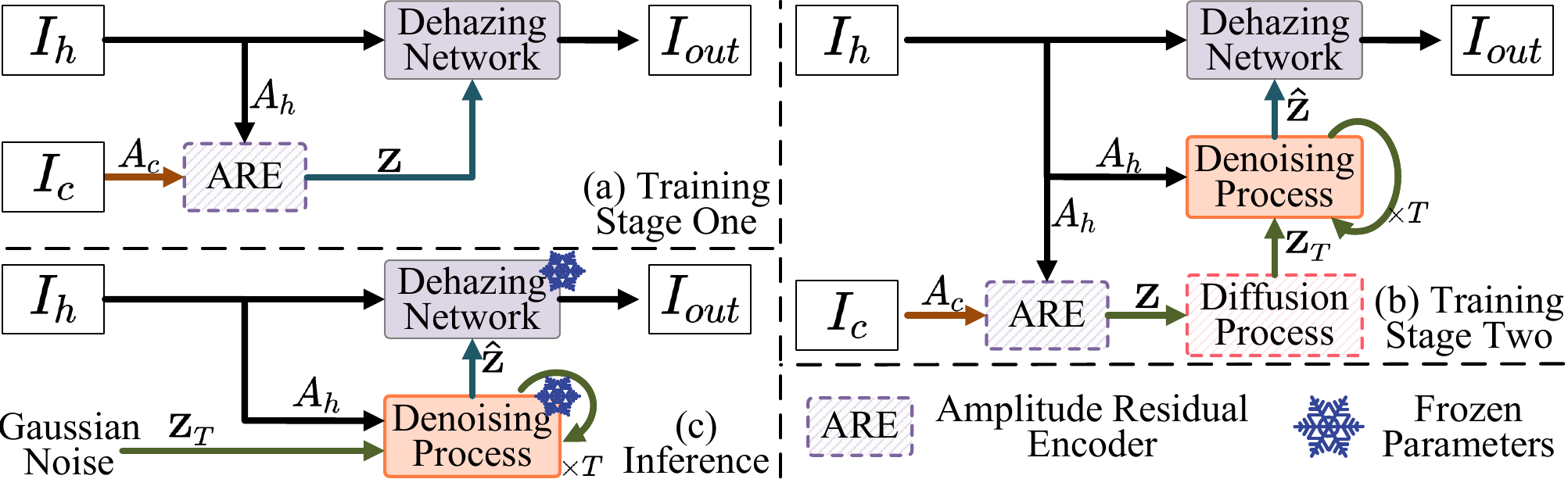}
    \vspace{-6mm}
   \caption{Schematic of training and inference strategies.}
   \label{fig:training}
   \vspace{-2mm}
\end{figure}

\vspace{1mm}
\noindent\textbf{Unsupervised learning Methods.}
Some pseudo-label-based methods~\cite{chen2021psd,wu2023ridcp,fang2025real,shao2020domain,chen2024prompt} attempt to explicitly model authentic haze characteristics during training. Nevertheless, these approaches not only introduce an additional re-hazing pipeline, but also heavily depend on domain expertise in haze genesis.
Dominant methods that train directly on unpaired data are mainly categorized into CycleGAN-like and contrastive learning-like methods. 
The CycleGAN-like methods~\cite{qiao2023learning,liu2020end,engin2018cycle,yang2022self} establish a cycle transformation framework of hazy and clear domains, where adversarial loss and cycle-consistency loss are used to supervise domain transfer and preserve content.
However, for different haze densities, the cycle-based framework between the two domains can lead to asymmetric optimization~\cite{park2020contrastive}.

The contrastive learning-like methods~\cite{park2020contrastive,chen2022unpaired,wang2024ucl,wang2024odcr} focus on constructing positive and negative sample pairs to maximize the mutual information between clean and hazy domains.  
Representatively, ODCR~\cite{wang2024odcr} projects image features into orthogonal space to reduce the impact of image content in contrast supervision. 
Unlike them, our motivation arises from the frequency domain properties of haze, \ie, reconstructing the amplitude spectrum that mainly manifests haze degradation, and removing haze by reconstructing the amplitude spectrum.

\subsection{Diffusion Model}
Recently, Diffusion Models (DMs)~\cite{ho2020denoising} have been developed to generate the desired content from Gaussian noise through a stochastic iterative denoising process.
In low-level vision, DMs serve as new generative models for reconstructing high-quality images in tasks such as super-resolution~\cite{xia2023diffir,kawar2022denoising,ye2024learning}, deblurring~\cite{chen2024hierarchical}, inpainting~\cite{lugmayr2022repaint}, and so on. 
Unlike paradigms that apply DMs directly for image synthesis~\cite{kawar2022denoising,lugmayr2022repaint}, the synthesis of prior representations can effectively avoid unpredictable artifacts and requires only a few iteration steps~\cite{xia2023diffir}.
Typically, HiDiff~\cite{chen2024hierarchical} and DiffMRI~\cite{li2024rethinking} propose to predict prior representations with DM to aid in deblurring and super-resolution, respectively.

However, they are limited to training with high-quality paired data and still require additional research to effectively handle unpaired input. 
In this work, we propose to learn the amplitude spectrum from unpaired data, enabling DMs to reconstruct them with lower computational costs.

\section{Methodology}
\label{sec:metho}

\subsection{Overall Architecture}
\label{FrDiff}

\begin{figure*}[t]
  \centering
    \includegraphics[width=1.0\linewidth,page=3]{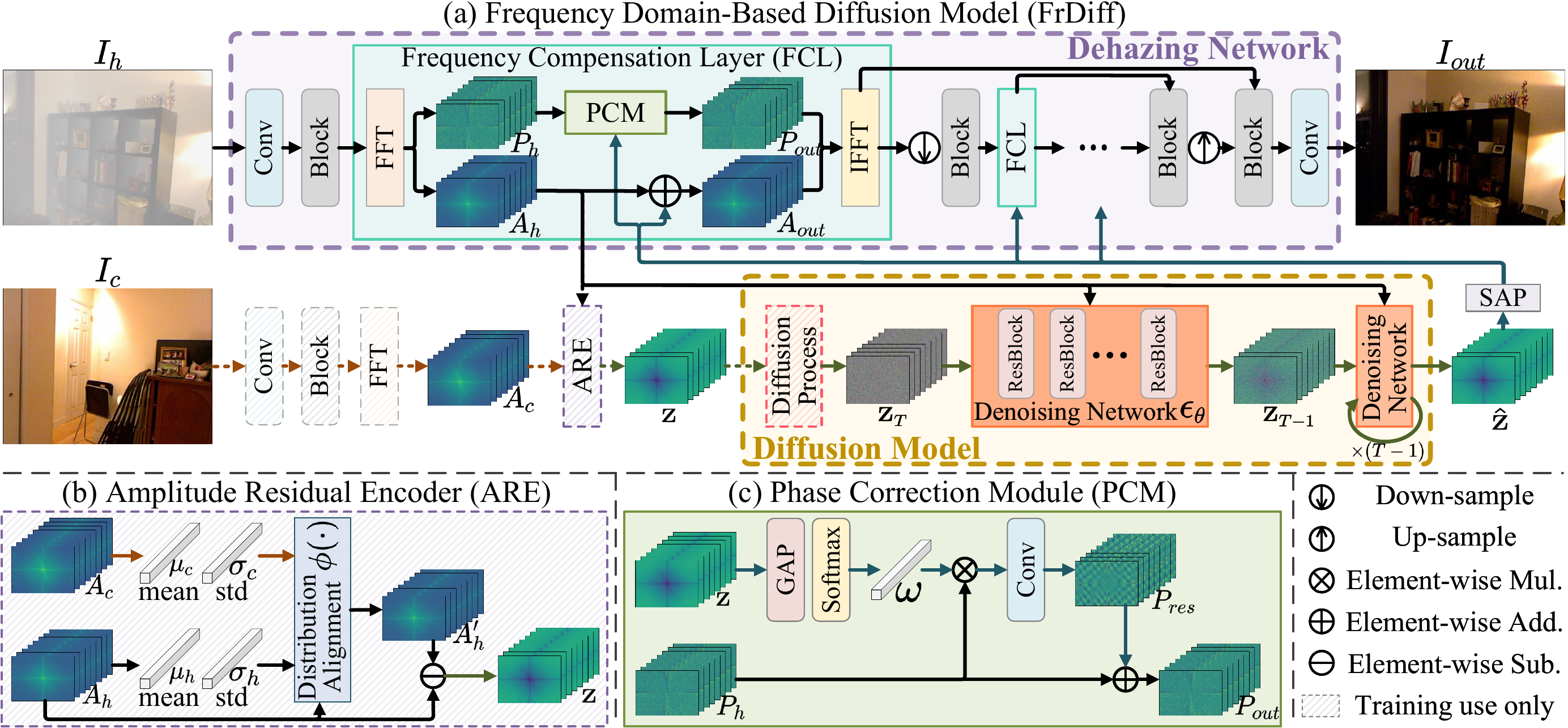}
    \vspace{-7mm}
   \caption{(a) Overview of \ours, which mainly consists of a dehazing network and a diffusion model (DM). 
   (b) Structure of the Amplitude Residual Encoder (ARE), which can yield amplitude residuals to compensate for hazy and clear domain gaps.
   (c) Structure of the Phase Correction Module (PCM), which can eliminate unwanted artifacts by refining the phase spectrum using an attention mechanism.
   }
   \label{fig:overview}
   \vspace{-2mm}
\end{figure*}

As shown in Fig.~\ref{fig:overview}(a), our \ours includes a dehazing network and a diffusion model (DM). 
Specifically, DM is employed to reconstruct the amplitude residuals, which represent the amplitude distribution gaps between the hazy and clear images. Meanwhile, at each scale of the dehazing network, we introduce a frequency compensation layer (FCL) that utilizes the DM's output to refine the frequency features. 
The core of \ours is to integrate the DM into the frequency dehazing process, so that DM learns useful knowledge from unpaired input and guides the dehazing network.

To efficiently optimize \ours, we follow the existing works~\cite{chen2024hierarchical,xia2023diffir} to adopt a two-stage training strategy. \ours takes the hazy image $I_h\in R^{3\times H\times W}$ and unpaired clear image $I_c\in R^{3\times H\times W}$ as inputs. 
As shown in~\cref{fig:training}(a), in stage one, we utilize the proposed amplitude residual encoder (ARE) to obtain amplitude residual $\mathrm{z}\in R^{C\times H'\times W'}$, and train the dehazing network for haze removal. 
In this case, the $\mathrm{z}$ is directly fed into the dehazing network without involving the diffusion and denoising processes. 
As shown in~\cref{fig:training}(b), in stage two, we joint train DM and dehazing network to reconstruct the amplitude residual for enhancing the dehazing process. In this stage, the $\mathrm{z}$ first adds noise to output $\mathrm{z}_T$ through the diffusion process, and then reconstructs the amplitude residual $\hat{\mathrm{z}}$ through multiple denoising processes. 
Next, we detail the two-stage training and inference processes in order.

\subsection{Stage One: Amplitude Residual Extraction}

In this stage, we aim to train a robust frequency-domain dehazing network by exploiting the amplitude residual $\mathrm{z}$ extracted from the unpaired input images.

Specifically, as shown in~\cref{fig:overview}(a), we adopt NAFNet~\cite{chen2022simple}, a simple UNet-based architecture, as our dehazing network and take the $I_h$ and $I_c$ as inputs. We perform the Fast Fourier Transform (FFT) on the features extracted through a convolutional layer and a basic block, obtaining the amplitude spectrum $A_h, A_c\in R^{C\times H'\times W'}$ of the hazy and clear image. Then, we use ARE to extract the corresponding amplitude residual $\mathrm{z}$. After that, we feed $\mathrm{z}$ into the FCL of the dehazing network to obtain the refined feature. More specifically, in FCL, we add $\mathrm{z}$ with the $A_h$ to get the refined amplitude spectrum $A_{out}$. Meanwhile, $\mathrm{z}$ is further fed into the proposed phase correction module (PCM) to refine the phase spectrum $P_{h}$ that is obtained from $I_h$. Finally, we use the Inverse Fast Fourier Transform (IFFT) to output the refined feature, and the dehazed image $I_{out}$ is outputted through the expanding path (down to up) of UNet. 
Notably, to further improve the model capacity, we also incorporate the FCL at multiple scales in the contracting path (up-to-down) of UNet and input $\mathrm{z}$ into these layers using spatial adaptive pooling (SAP). Please refer to the supplementary for the detailed structure.
Following, we discuss the core components, ARE and PCM.

\vspace{1mm}
\noindent\textbf{Amplitude Residual Encoder.}
The purpose of the ARE is to obtain an amplitude residual that compensates for the gap in distribution between hazy and clear images without altering the texture structures present in the hazy input.

In detail, as shown in~\cref{fig:overview}(b), ARE takes the amplitude spectrum $A_h, A_c$ of the hazy and clear image as inputs. 
We first compute the mean $\mu_h,\mu_c$ and standard deviation $\sigma_h,\sigma_c$ to characterize their distributions, formulated as:
{\small\begin{align}
    &\mu_{\{h,c\}}\! = \!\frac{1}{C\!\cdot\! H'\!\cdot\! W'}\sum_{i=1}^{C}\sum_{j=1}^{H'}\sum_{k=1}^{W'}A_{\{h,c\}}(i,j,k),\\
    \sigma_{\{h,c\}}\!\!&=\!\!{\big(}\frac{1}{C\!\!\cdot\!\! H'\!\!\cdot\!\! W'}\!\!\sum_{i=1}^{C}\!\sum_{j=1}^{H'}\!\sum_{k=1}^{W'}(A_{\{h,c\}}\!(i,j,k)\!\!-\!\!\mu_{\{h,c\}})^2{\big)} ^{1/2}.
\end{align}}
Then, we align the amplitude distributions of the hazy image to those of the clear image using a defined distribution alignment function $\phi(\cdot)$. This process can be formulated as:
\begin{equation}
    \begin{aligned}
        A_h' &= \phi (A_h;\mu_h,\sigma_h;\mu_c,\sigma_c)\\
        &= \frac{\sigma_c}{\sigma_h} (A_h-\mu_h)+\mu_c
    \end{aligned},
    \label{eq:daf}
\end{equation}
where $A_h'$ is the aligned amplitude spectrum, consistent with the distribution of the unpaired clear data.
Such practice not only preserves the texture structure of the haze input but also does not introduce learnable model parameters. 
Finally, the output can be obtained by $\mathrm{z} = A_h'\ominus A_h$,
where $\ominus$ denotes the element-wise subtraction. Haze removal can be achieved by using $\mathrm{z}$ to compensate for the frequency domain features in the dehazing network. 

To minimize the complexity, we use one ARE to obtain the amplitude residuals which are fed into multiple FCLs. SAP is used for adaptive pooling according to different scales.
Compared to directly outputting the entire amplitude spectrum and image for reconstruction, outputting the amplitude residual $\mathrm{z}$ for reconstruction can effectively reduce the burdens of the DM and requires fewer iterations step. We provide detailed analysis in~\cref{Iterations}.

\vspace{1mm}
\noindent\textbf{Phase Correction Module.} 
Compared to the amplitude spectrum, the phase spectrum manifests more texture structure and is highly immune to illumination contrast and color shifts~\cite{skarbnik2009importance,oppenheim1981importance}. However, we found that severe haze can obscure the texture structure, resulting in unrealistic textures in the dehazed image as shown in~\cref{fig:case_ablation}. Therefore, we introduce PCM to refine the phase spectrum further.

In detail, as shown in~\cref{fig:overview}(c), PCM takes the phase spectrum $P_h$ and the amplitude residual $\mathrm{z}$ obtained from ARE as inputs. We use $\mathrm{z}$ to obtain the weight vector $\omega\in R^{C\times 1\times 1}$, and then generate the phase residual $P_{res}\in R^{C\times H'\times W'}$ through a convolutional layer, formulated as:
\begin{align}
    &P_{res} = \mathrm{Conv}(\omega\otimes P_h),\\
    &\omega = \mathrm{SoftMax}(\mathrm{GAP}(\mathrm{z})),
\end{align}
where $\mathrm{GAP}(\cdot)$ and $\mathrm{SoftMax}(\cdot)$ are the global average pooling and softmax, respectively.
$\mathrm{Conv}(\cdot)$ and $\otimes$ are the convolutional layer and element-wise multiplication, respectively.
Finally, the refined phase spectrum can be obtained by $P_{out} = P_{h}\oplus P_{res}$, where $\oplus$ is element-wise addition.

This design incorporates haze intensity information from the amplitude to generate phase residuals with minimal cost, making it significantly simpler than the typical self-attention~\cite{vaswani2017attention}. We provide detailed analysis in~\cref{pcm}.

\vspace{1mm}
\noindent\textbf{Optimization Objective.}
It is infeasible to supervise model training with the strict pixel-level loss function under the unpaired data setting. Therefore, we follow existing works~\cite{park2020contrastive,wang2024odcr,wang2024ucl}, using the same adversarial loss $\mathcal{L}_{GAN}$ and patchwise contrast loss $\mathcal{L}_{PatchNCE}$ to encourage the dehazed results have the same domain distribution as the clear images, formulated as:
\begin{equation}
\mathcal{L}_{s1}=\lambda_{GAN}\mathcal{L}_{GAN}+\lambda_{PatchNCE}\mathcal{L}_{PatchNCE},
\end{equation}
where the hyper-parameters $\lambda_{GAN}$ and $\lambda_{PatchNCE}$ determine the relative importance of each term. 

\subsection{Stage Two: DM-based Reconstruction}
In this stage, we aim to train DM to learn the reconstruction of amplitude residual from Gaussian noise, which enhances the capability of the dehazing network through FCL.
As shown in~\cref{fig:overview}(a), following the conditional denoising diffusion probabilistic models~\cite{chen2024hierarchical,li2024rethinking,xia2023diffir}, our DM consists of the forward diffusion process and the reverse denoising process.
In the forward diffusion process, we first adopt the proposed ARE to generate the ground truth amplitude residual $\mathrm{z}$. 
Then, we add noise to $\mathrm{z}$ by a diffusion process and get the $\mathrm{z}_T\in R^{C\times H'\times W'}$ with the same distribution as the Gaussian noise.
In the reverse denoising process, we use the denoise network to generate the reconstructed amplitude residual $\hat{\mathrm{z}}\in R^{C\times H'\times W'}$ using the extracted amplitude spectrum $A_h$ as the condition and $\mathrm{z}$ as the target. Following, we describe the diffusion and denoising process in detail.

\vspace{1mm}
\noindent\textbf{Diffusion Process.}
Same as the existing works~\cite{ho2020denoising,song2021denoising}, we perform the forward Markov process starting from $\mathrm{z}$, and gradually add Gaussian noise by $T$ iterations as follows:
\begin{equation}
    q(\mathrm{z}_{T}\mid \mathrm{z})=\mathcal{N} (\mathrm{z}_{T} ;\sqrt{\bar{\alpha}_{T}}\mathrm{z},(1-\bar{\alpha}_{T})\mathrm{I}),
\end{equation}
where $T$ is the total number of iteration steps. $\mathcal{N}(\cdot)$ denotes the Gaussian distribution. $\alpha=1-\beta_t$, $\bar{\alpha}_t=\prod_{i=1}^t \alpha_i$, where $t\in\{1,\dots,T\}$, $\beta_{1:T} \in (0,1)$, are hyperparameters derived through iterative derivation with reparameterization~\cite{kingma2013auto} to control the amount of noise added at each step.

\setlength{\tabcolsep}{1.8mm}{
\begin{table*}[t]\small
\centering
\begin{tabular}{clcccccccccc}
\toprule
\multicolumn{2}{c}{\multirow{2}{*}{Methods}}       & \multicolumn{2}{c}{SOTS-Indoor~\cite{li2018benchmarking}} & \multicolumn{2}{c}{SOTS-Outdoor~\cite{li2018benchmarking}} & \multicolumn{2}{c}{HSTS-Synth~\cite{li2018benchmarking}}& \multicolumn{2}{c}{I-HAZE~\cite{ancuti2018haze}}   &  \multicolumn{2}{c}{Efficiency} \\
\cmidrule[0.1pt](lr{0.125em}){3-4}\cmidrule[0.1pt](lr{0.125em}){5-6}\cmidrule[0.1pt](lr{0.125em}){7-8}\cmidrule[0.1pt](lr{0.125em}){9-10}\cmidrule[0.1pt](lr{0.125em}){11-12}
 &  &  \cellcolor{LightGray}PSNR & \cellcolor{LightGray}SSIM & \cellcolor{LightGray}PSNR & \cellcolor{LightGray}SSIM    & \cellcolor{LightGray}PSNR & \cellcolor{LightGray}SSIM& \cellcolor{LightGray}PSNR & \cellcolor{LightGray}SSIM   & \cellcolor{LightGray}\#Param(M) & \cellcolor{LightGray}FLOPs(G) \\
\midrule
\multirow{4}{*}{\makecell[c]{\rotatebox{90}{Supervised}}} 
 &   EPDN~\cite{qu2019enhanced}    & 25.06 &0.931 & 20.47 &0.896 & 20.37 & 0.877& 15.02 & 0.763 & 17.38& 4.826    \\
 &   \cellcolor{LightGray}AOD-Net~\cite{li2017aod}    &  \cellcolor{LightGray}19.06& \cellcolor{LightGray}0.850&\cellcolor{LightGray}20.08 &\cellcolor{LightGray}0.861 & \cellcolor{LightGray}20.55 & \cellcolor{LightGray}0.897 &\cellcolor{LightGray}- & \cellcolor{LightGray}- & \cellcolor{LightGray}-& \cellcolor{LightGray}-   \\
  &   MSCNN~\cite{ren2016single}    & 19.84 &0.833 & 14.62 & \secondnum{0.908} & - & - & - & - & 0.01 & -   \\
 &   \cellcolor{LightGray}FFANet~\cite{qin2020ffa}    &  \cellcolor{LightGray}\bestnum{36.36}& \cellcolor{LightGray}\bestnum{0.993}&\cellcolor{LightGray}20.23 &\cellcolor{LightGray}0.905 & \cellcolor{LightGray}- & \cellcolor{LightGray}- &\cellcolor{LightGray}12.00 & \cellcolor{LightGray}0.592 & \cellcolor{LightGray}4.456 &\cellcolor{LightGray}288.3   \\
\midrule
\multirow{11}{*}{\makecell[c]{\rotatebox{90}{Unsupervised}}} &  DCP~\cite{he2010single}    &   13.10 & 0.699 & 19.13&  0.815 & 14.84 &0.761   &  13.10 & 0.699 &- & -  \\
&  \cellcolor{LightGray}CycleGAN~\cite{zhu2017unpaired}    &     \cellcolor{LightGray}21.34&  \cellcolor{LightGray}0.898 &  \cellcolor{LightGray}20.55 & \cellcolor{LightGray}0.856 & \cellcolor{LightGray}18.52& \cellcolor{LightGray}0.831  &   \cellcolor{LightGray}15.29 & \cellcolor{LightGray}0.756 &  \cellcolor{LightGray}11.38 & \cellcolor{LightGray}56.89  \\
&  YOLY~\cite{li2021you}    &     15.84 & 0.819 &  14.75&  0.857 & 21.02 & 0.905 &   14.74 & 0.688 &  39.99 &  - \\
&  \cellcolor{LightGray}CUT~\cite{park2020contrastive}    & \cellcolor{LightGray}21.11  &\cellcolor{LightGray}0.880   & \cellcolor{LightGray}17.81  & \cellcolor{LightGray}0.713  &  \cellcolor{LightGray}17.16 & \cellcolor{LightGray}0.715  &  \cellcolor{LightGray}14.76 & \cellcolor{LightGray}0.694  &   \cellcolor{LightGray}11.38 & \cellcolor{LightGray}56.89  \\
&  RefineDNet~\cite{zhao2021refinednet}    &     24.36&  0.939 &   19.84 & 0.853 & 21.69 &0.904  &   13.60 &  0.660 &  65.80 & 75.41 \\
&   \cellcolor{LightGray}PSD~\cite{chen2021psd}    &  \cellcolor{LightGray}15.02 & \cellcolor{LightGray}0.764 & \cellcolor{LightGray}15.63 & \cellcolor{LightGray}0.834 & \cellcolor{LightGray}19.37 & \cellcolor{LightGray}0.824 & \cellcolor{LightGray}15.30 & \cellcolor{LightGray}0.752 & \cellcolor{LightGray}33.11& \cellcolor{LightGray}182.5  \\
&  RIDCP~\cite{wu2023ridcp}    &    18.36 & 0.756 & 21.72 &0.839& 22.77 & 0.840&   15.07 & 0.706&  28.72 & - \\
&  \cellcolor{LightGray}CORUN~\cite{fang2025real}    &   \cellcolor{LightGray}19.65 & \cellcolor{LightGray}0.742 & \cellcolor{LightGray}\secondnum{22.11} &\cellcolor{LightGray}0.869& \cellcolor{LightGray}22.73 & \cellcolor{LightGray}0.882&   \cellcolor{LightGray}15.22 & \cellcolor{LightGray}0.749&  \cellcolor{LightGray}- & \cellcolor{LightGray}- \\
&  PTTD~\cite{chen2024prompt}    &    19.37 & 0.755 & 21.76 &0.862 & 22.47 & 0.899&   15.34 & 0.755 &  2.611 & - \\
&  \cellcolor{LightGray}CDD-GAN~\cite{chen2022unpaired}   &   \cellcolor{LightGray}24.61   & \cellcolor{LightGray}0.918   &  \cellcolor{LightGray}20.82& \cellcolor{LightGray}0.841 & \cellcolor{LightGray}22.16 & \cellcolor{LightGray}\secondnum{0.911}     & \cellcolor{LightGray}14.65&  \cellcolor{LightGray}0.672   & \cellcolor{LightGray}11.38 & \cellcolor{LightGray}56.89 \\
&  $D^4$~\cite{yang2022self}    &    25.42 & 0.932 & 20.96 & 0.859& \secondnum{22.84} & \bestnum{0.923} &   \bestnum{15.61} & \bestnum{0.780}&  10.70 & 2.246 \\
&  \cellcolor{LightGray}UCL-Dehaze~\cite{wang2024ucl}   & \cellcolor{LightGray}24.53     &  \cellcolor{LightGray}0.915  & \cellcolor{LightGray}20.14 & \cellcolor{LightGray}0.789 & \cellcolor{LightGray}20.38 & \cellcolor{LightGray}0.791     & \cellcolor{LightGray}14.65& \cellcolor{LightGray}0.672    &  \cellcolor{LightGray}11.38 & \cellcolor{LightGray}56.89 \\
&  ODCR~\cite{wang2024odcr}    &    {26.32} & {0.945} & 18.94 & 0.835& 18.29 & 0.810 &   14.95 & 0.756&  11.38 & 56.89 \\
 &   \cellcolor{LightGray}\ours    &   \cellcolor{LightGray}\secondnum{27.43}   & \cellcolor{LightGray}\secondnum{0.957}   & \cellcolor{LightGray}\bestnum{22.75} & \cellcolor{LightGray}\bestnum{0.914} & \cellcolor{LightGray}\bestnum{23.24} & \cellcolor{LightGray}\bestnum{0.923}  &  \cellcolor{LightGray}\secondnum{15.35}& \cellcolor{LightGray}\secondnum{0.763}    &\cellcolor{LightGray}8.76 &  \cellcolor{LightGray}54.56  \\  
\bottomrule
\end{tabular}
\vspace{-2mm}
\caption{Quantitative comparison on the SOTS-Indoor~\cite{li2018benchmarking}, SOTS-Outdoor~\cite{li2018benchmarking}, HSTS-Synth~\cite{li2018benchmarking}, and I-HAZE~\cite{ancuti2018haze} datasets. 
FLOPs(G) is computed on images with the size of $256\times256$. \bestnum{Red} and \secondnum{blue} indicate the best and the second best performance, respectively.}
\vspace{-3mm}
\label{tab:tab1}
\end{table*}}

\vspace{1mm}
\noindent\textbf{Denoising process.}
Aiming to reconstruct the $\hat{\mathrm{z}}$ from Gaussian noise, we perform the Markov chain that runs backward from $\mathrm{z}_T$ to $\mathrm{z}$, and gradually remove the noise by $T$ iterations.
In the inverse step from $\mathrm{z}_t$ to $\mathrm{z}_{t-1}$:
\begin{align}
q(\mathrm{z}_{t-1} \mid \mathrm{z}_t, \mathrm{z})=\mathcal{N}&(\mathrm{z}_{t-1};\mu_t(\mathrm{z}_t, \mathrm{z}), \frac{1-\bar{\alpha}_{t-1}}{1-\bar{\alpha}_t} \beta_t \mathrm{I}),\label{1}\\
\mu_t(\mathrm{z}_t, \mathrm{z})&=\frac{1}{\sqrt{\alpha_t}}(\mathrm{z}_t-\frac{1-\alpha_t}{\sqrt{1-\bar{\alpha}_t}} \epsilon),
\label{2}
\end{align}
where $\epsilon$ represents the noise in $\mathrm{z}_t$, which is the only uncertain variable that needs to be estimated at each step using the denoising network. Therefore, we use a neural network consisting of a series of stacked ResBlocks~\cite{he2016deep}, denoted as $\epsilon_\theta$, to estimate the noise with the extracted amplitude spectrum $A_h$ as the condition. Then, we further substitute $\epsilon_\theta$ into Eqs.~(\ref{1}) and (\ref{2}) to get:
\begin{equation}
\mathrm{z}_{t-1}\!=\!\frac{1}{\sqrt{\alpha_t}}(\mathrm{z}_t-\frac{1-\alpha_t}{\sqrt{1-\bar{\alpha}_t}} \epsilon_\theta(\mathrm{z}_t, A_h, t))\!+\!\sqrt{1\!-\!\alpha_t} \epsilon_t,
\label{3}
\end{equation}
where $\epsilon_t\!\sim\!\mathcal{N}(0,\mathrm{I})$. $\epsilon_\theta(\mathrm{z}_t, A_h, t)$ is the noise estimated by the denoising network. By repeating the $T$ times sampling iterations in Eq.~(\ref{3}), we can obtain the reconstructed amplitude residual $\hat{\mathrm{z}}$. The purpose of using ResBlocks as the denoising network is to ensure the same resolution of inputs and outputs while minimizing the model parameters. As shown in~\cref{fig:overview}(a), the output $\hat{\mathrm{z}}$ from the DM is finally utilized to enhance the dehazing process of the FCL.

\vspace{1mm}
\noindent\textbf{Optimization Objective.}
In this stage, our objective is to joint train the DM,~\ie the denoising network $\epsilon_\theta$, and whole dehazing network. To achieve this, we additionally include the diffusion loss $\mathcal{L}_{diff}$ based on $\mathcal{L}_{s1}$, formulated as:
\begin{equation}
\mathcal{L}_{s2}=\mathcal{L}_{s1} + \lambda_{diff}\mathcal{L}_{diff}, \ \   \mathcal{L}_{diff}=\left \| \mathrm{z}-\hat{\mathrm{z}} \right \| _1.
\label{loss2}
\end{equation}
\subsection{Inference}
\label{inference}
To enable the output of DM can compensate for the amplitude of hazy input, we use the amplitude residuals $\mathrm{z}$ from ARE as GT and the amplitude of hazy input as the condition when DM training.
During inference, as a probabilistic generative model, DM can construct the required data samples from Gaussian noise through a stochastic iterative denoising process~\cite{ho2020denoising,xia2023diffir}. 
As shown in~\cref{fig:training}(c), given a hazy input image $I_h$, we first extract the amplitude spectrum $A_h$ from the $I_h$ as the condition of DM. Then, we randomly sample pure Gaussian noise $\mathrm{z}_T$. After $T$ times denoising process in~\cref{3}, DM generates the amplitude residual $\hat{\mathrm{z}}$ using $\mathrm{z}_T$ and $A_h$. Finally, we feed $\hat{\mathrm{z}}$ to the dehazing network, which includes FCL, to achieve the dehazed result.

\section{Experiments}
\label{sec:exper}
\subsection{Datasets and Metrics}
We evaluate our method on widely-used RESIDE~\cite{li2018benchmarking}, I-HAZE~\cite{ancuti2018haze}, and Fattal's~\cite{fattal2014dehazing} datasets, which cover synthetic, artificial, and real-world images. 
Specifically, the RESIDE~\cite{li2018benchmarking} dataset contains several subsets: (a) ITS, which contains 13,990 synthetic indoor hazy-clear pairs. (b) SOTS-Indoor/Outdoor, which contains 500 synthetic indoor/outdoor hazy-clear pairs. (c) HSTS-Synth/Real, which contains 10 synthetic hazy-clear pairs and 10 real-world hazy images without ground-truth images, respectively. (d) URHI, which contains over 4,000 real hazy images without ground-truth images. I-HAZE~\cite{ancuti2018haze} dataset contains 35 artificial hazy-clear pairs produced by professional haze generators. Fattal's~\cite{fattal2014dehazing} dataset includes 31 real-world hazy images in various scenes. For fair comparisons, we follow the previous works~\cite{yang2022self,wang2024odcr} to use ITS~\cite{li2018benchmarking} as the training set and the rest as the test set. We keep the same evaluation metrics of PSNR(dB) and SSIM as previous works~\cite{yang2022self,wang2024odcr}.

\subsection{Implementation Details}
In the dehazing network, we configure the number of basic blocks as [4, 4, 6, 10], and the number of channels $C$ as 64. In the denoising network, we set the ResBlock number to 5 and the iteration step $T$ to 8. The $\lambda_{GAN}$, $\lambda_{PatchNCE}$, and $\lambda_{diff}$ are all set to 1. We use Adam~\cite{kingma2014adam} optimizer with $\beta_{1}=0.9$ and $\beta_{2}=0.999$, learning rate is $1\times 10^{-4}$. The epoch number for each stage is 200. We set the batch size as $8$ and the input patch size as $256\times 256$ and augment the data with random horizontal and vertical flips. 
All experiments are performed on an NVIDIA RTX A6000 GPU.

\setlength{\tabcolsep}{0.4mm}{
\begin{table}[t]\small
\centering
\begin{tabular}{l|ccc}
\toprule
Methods       & HSTS-Real\cite{li2018benchmarking} & Fattal's\cite{fattal2014dehazing} & URHI\cite{li2018benchmarking}\\
\midrule
\myrowcolour%
CycleGAN\cite{zhu2017unpaired} & 0.9286/30.021 & 0.3440/20.671 & 1.1669/\bestnum{27.057}\\
CUT~\cite{park2020contrastive} & 1.3661/38.992 & 0.4330/30.112 &1.4627/36.671\\
\myrowcolour%
RefineDNet\cite{zhao2021refinednet} & 0.8948/31.608 & 0.3996/21.479 & 0.9926/29.788\\
CDD-GAN\cite{chen2022unpaired} & 1.0744/35.762 & 0.3942/31.260 & 1.2834/35.234\\
\myrowcolour%
$D^4$\cite{yang2022self} & 1.1298/35.397 & 0.4221/20.330 & 1.2776/34.360\\
PSD\cite{chen2021psd}& 1.2932/34.271 & 0.3719/19.326 & 1.2467/31.256\\
\myrowcolour%
RIDCP\cite{wu2023ridcp} & \secondnum{0.8869}/\secondnum{28.146} & \secondnum{0.3182}/21.872 & 0.9117/33.668\\
CORUN\cite{fang2025real} & 0.9266/31.265 & 0.3370/20.913 & \bestnum{0.8642}/31.952\\
\myrowcolour%
PTTD\cite{chen2024prompt} & 0.9142/30.695 & 0.3874/19.255 & 0.8943/\secondnum{28.772}\\
UCL-Dehaze\cite{wang2024ucl} & 0.8917/31.217 & 0.4117/\secondnum{18.849} & 1.1428/33.073\\
\myrowcolour%
ODCR\cite{wang2024odcr} & 1.2410/35.800 & 0.3871/30.698 & 1.2734/35.601\\
\midrule
\ours & \bestnum{0.8732}/\bestnum{26.972} & \bestnum{0.2435}/\bestnum{18.747} & \secondnum{0.8813}/32.022\\
\bottomrule
\end{tabular}
\vspace{-2mm}
\caption{Quantitative comparison (FADE$\downarrow$ and BRISQUE$\downarrow$) on HSTS-Real~\cite{li2018benchmarking}, Fattal's~\cite{fattal2014dehazing}, and URHI~\cite{li2018benchmarking} datasets.}
\label{tab:real}
\vspace{-4mm}
\end{table}}

\begin{figure*}[t]
  \centering
    \includegraphics[width=0.95\linewidth,page=5]{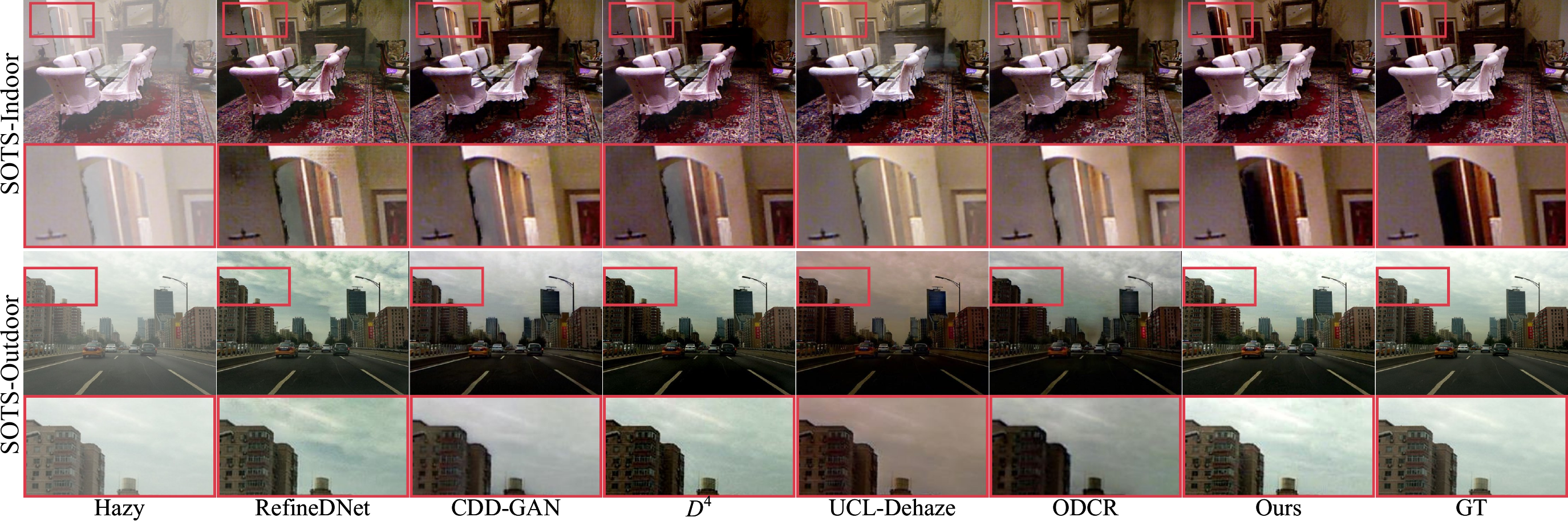}
    \vspace{-3mm}
   \caption{Visual results on SOTS-Indoor~\cite{li2018benchmarking} and SOTS-Outdoor~\cite{li2018benchmarking} datasets. 
   Zoom in to see better visualization.}
   \label{fig:case_synth}
   \vspace{-2mm}
\end{figure*}

\begin{figure*}[t]
  \centering
    \includegraphics[width=0.89\linewidth,page=6]{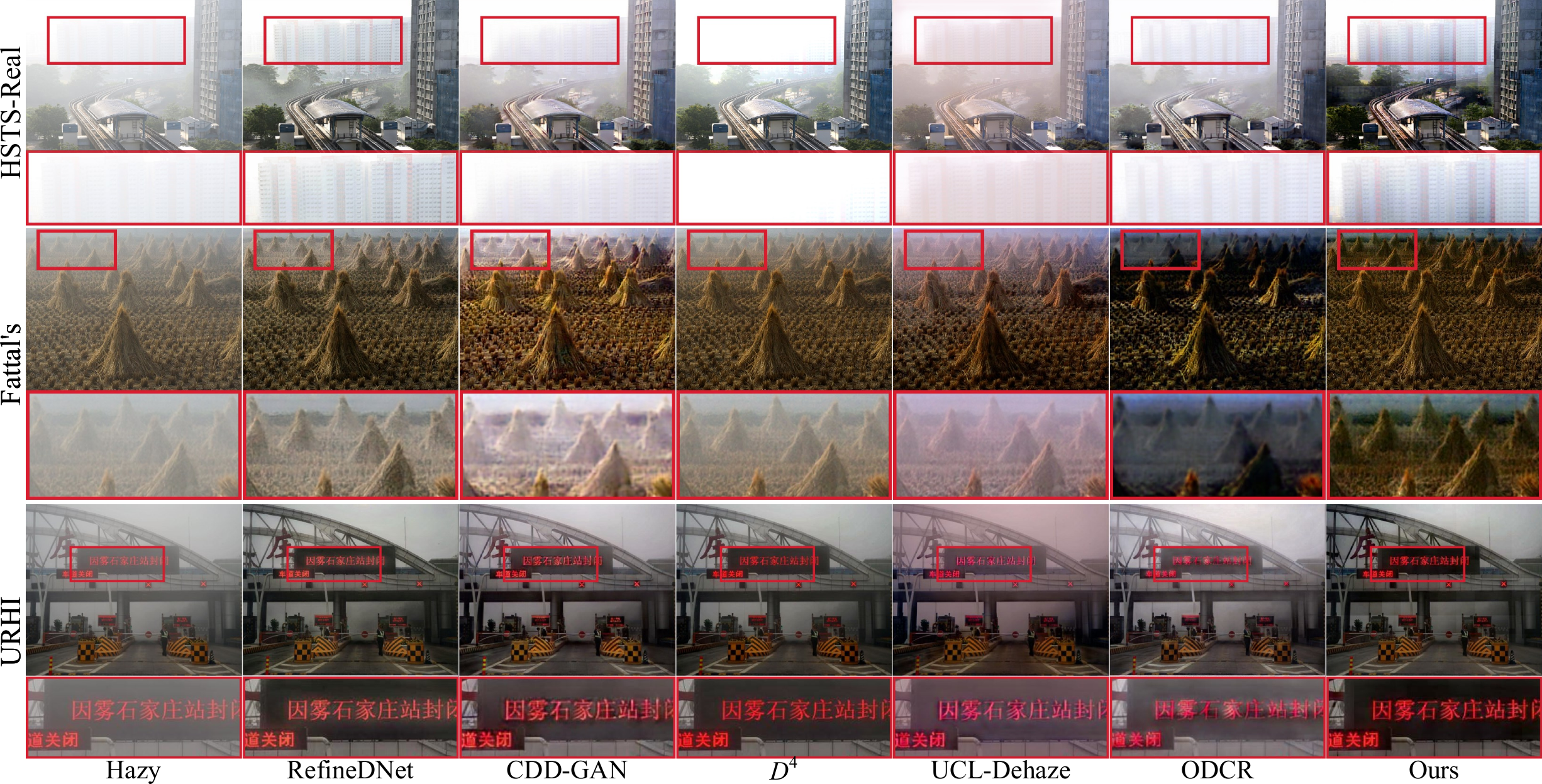}
     \vspace{-3mm}
   \caption{Visual results on real-world HSTS-Real~\cite{li2018benchmarking}, Fattal's~\cite{fattal2014dehazing}, and URHI~\cite{li2018benchmarking} datasets. 
   Zoom in to see better visualization.}
   \label{fig:case_real}
   \vspace{-2mm}
\end{figure*}

\subsection{Comparisons with State-of-the-art Methods}
We evaluate the proposed method against SOTA models, which consist of supervised methods~\cite{qu2019enhanced,li2017aod,qin2020ffa,ren2016single} and unsupervised methods~\cite{he2010single,zhu2017unpaired,li2021you,park2020contrastive,zhao2021refinednet,chen2022unpaired,yang2022self,wang2024ucl,wang2024odcr,wu2023ridcp,chen2021psd,fang2025real,chen2024prompt}.
For fair comparisons, we obtain the performance from original paper or reproduce results by officially codes.

\begin{figure*}[t]
  \begin{minipage}[h]{0.6\textwidth}\small
        \setlength{\tabcolsep}{0.6mm}{
        \begin{tabular}{l|>{\centering\arraybackslash}p{1.2cm}>{\centering\arraybackslash}p{0.7cm}>{\centering\arraybackslash}p{0.9cm}>{\centering\arraybackslash}p{1.6cm}>{\centering\arraybackslash}p{1.9cm}|cc}
        \toprule
        \multirow{2}{*}{Methods}       & \multicolumn{5}{c|}{Components} &   \multicolumn{2}{c}{Metrics} \\
        \cmidrule[0.1pt](lr{0.125em}){2-6}\cmidrule[0.1pt](lr{0.125em}){7-8}
        \myrowcolour%
         & Diffusion & ARE & PCM  & Multi-Scale  &  Joint-Training  & PSNR & SSIM   \\
        \midrule
        \textit{w/o} DM   &    \XSolidBrush  & \Checkmark & \Checkmark  & \Checkmark  &  \XSolidBrush & 24.66 & 0.922   \\
        \myrowcolour%
        \textit{w/o} ARE   &  \Checkmark  & \XSolidBrush  & \Checkmark   & \Checkmark  &  \Checkmark  & 27.12 & 0.943   \\
        \textit{w/o} PCM   &    \Checkmark  & \Checkmark & \XSolidBrush   & \Checkmark &  \Checkmark   & 27.14 & 0.952   \\
        \myrowcolour%
        \textit{w/o} MS   &    \Checkmark  & \Checkmark & \Checkmark    & \XSolidBrush &  \Checkmark  &26.80 & 0.950   \\
        \textit{w/o} JT   &    \Checkmark  & \Checkmark & \Checkmark   & \Checkmark &  \XSolidBrush  & 26.77 & 0.950   \\
        \myrowcolour%
        Full model  &    \Checkmark  & \Checkmark & \Checkmark   & \Checkmark &  \Checkmark   & \textbf{27.43} & \textbf{0.957}   \\
        \bottomrule
        \end{tabular}
        \vspace{-2mm}
        \captionof{table}{Ablation study of each component on the SOTS-Indoor~\cite{li2018benchmarking} dataset.}
        \label{tab:ablation}}
  \end{minipage}%
  \begin{minipage}[h]{0.4\textwidth}
      \centering
        \includegraphics[width=1.0\linewidth,page=7]{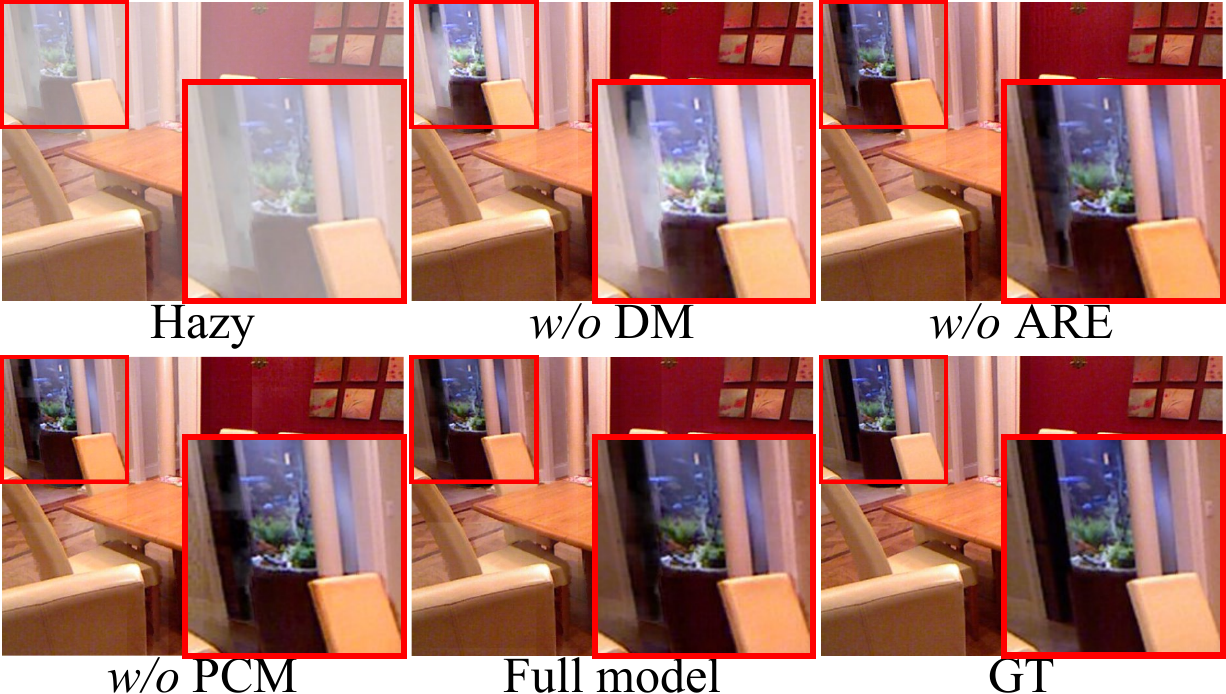}
      \vspace{-6mm}
       \caption{Visualization of ablation study on \ours.}
       \label{fig:case_ablation}
  \end{minipage}
  \vspace{-4mm}
\end{figure*}

\vspace{1mm}
\noindent\textbf{Results on Synthetic Datasets.}
\cref{tab:tab1} shows that \ours performs well on SOTS-Indoor compared to other unsupervised-based methods.
Contrastive learning-like methods (\eg,~UCL-Dehaze~\cite{wang2024ucl}, ODCR~\cite{wang2024odcr}) tend to produce sub-optimal results because they implicitly incorporate haze-unrelated content information.
Unlike them, \ours is inspired by haze properties in the frequency domain and removes haze by enabling the DM to learn the amplitude reconstruction of clear images. 
Due to such merits, \ours also performs favorably against other methods with less number of parameters and comparable FLOPs in other test datasets with more complex scenarios (\eg,~SOTS-Outdoor~\cite{li2018benchmarking}, HSTS-Synth~\cite{li2018benchmarking}).
This demonstrates the powerful haze removal capability of our \ours.

To further compare the visual quality of different methods, we present their dehazed results in Fig.~\ref{fig:case_synth}. The results of \ours have better visual quality. For example, in the first case, \ours removes the thick haze inside the door, while other methods still observe the haze. In the second case, \ours can recover the proper illumination contrast and sky color. The results verify that \ours can reconstruct the amplitude spectrum which assists in producing clearer results.

\begin{figure}[t]
  \begin{minipage}[h]{0.24\textwidth}\small
    \centering
        \setlength{\tabcolsep}{0.2mm}{
        \centering
        \begin{tabular}{l|cc}
        \toprule
        Methods    & PSNR  & SSIM   \\
        \midrule
        \myrowcolour%
        Self-Attn~\cite{dosovitskiy2020image} & 27.27 & 0.952 \\
        ARE(RGB Space) &  25.16&0.920 \\
        \midrule  
        \myrowcolour%
         ARE(\textit{w/} mean) &   27.36 &  0.955 \\
         ARE(\textit{w/} std) &   27.23 &  0.951 \\
         \myrowcolour%
         ARE(\textit{w/} mean+std) &   \textbf{27.43} &  \textbf{0.957}\\
        \bottomrule
        \end{tabular}
        \vspace{-2mm}
        \captionof{table}{Ablation study on ARE.}
        \label{tab:ae}
        }
  \end{minipage}%
  \begin{minipage}[h]{0.24\textwidth}\small
    \centering
    \setlength{\tabcolsep}{0.2mm}{
    \begin{tabular}{l|cc}
    \toprule
    Methods    & PSNR  & SSIM   \\
    \midrule
    \myrowcolour%
    Self-Attn\cite{dosovitskiy2020image} & 27.21  & 0.952\\
    Spatial Attn\cite{jaderberg2015spatial} & 27.16  & 0.952 \\
    \myrowcolour%
    Channel Attn\cite{hu2018squeeze} & 27.29  &0.954 \\
    Mixed Attn\cite{woo2018cbam}& 27.38  & 0.956 \\
    \midrule
    \myrowcolour%
    PCM &   \textbf{27.43} &  \textbf{0.957} \\
    \bottomrule
    \end{tabular}
    \vspace{-1.5mm}
    \captionof{table}{Ablation study on PCM.}
    \label{tab:pcm}
    }
  \end{minipage}
  \vspace{-3mm}
\end{figure}

\vspace{1mm}
\noindent\textbf{Results on Real-World Datasets.}
We evaluate \ours performance in real-world scenarios by comparing on the HSTS-Real~\cite{li2018benchmarking}, Fattal's~\cite{fattal2014dehazing}, and URHI~\cite{li2018benchmarking} datasets.
Since there is no corresponding ground truth, we employ two widely-used NR-IQA metrics FADE~\cite{choi2015referenceless} and BRISQUE~\cite{mittal2012no}. \cref{tab:real} shows that \ours achieves the highest performance on HSTS-Real~\cite{li2018benchmarking} and Fattal’s~\cite{ fattal2014dehazing} dataset, and the favourable performance on URHI~\cite{li2018benchmarking}. Verifies our generalizability to handle complex hazes.

In addition, we also compare the visual quality of different methods in~\cref{fig:case_real}. In the first and second cases, \ours is able to remove the heavy haze in more distant places. In the third case, \ours recovers clear text affected by haze. Experimental results demonstrate the robustness of \ours in realistic haze scenarios, in particular the ability to produce more natural haze-free images with perceptually pleasant and consistent quality. Please refer to the supplementary for more quantitative and qualitative results.

\subsection{Ablation Study}
In this section, we study the effectiveness of each component.
For fair comparisons, all experiments are performed in the same settings and evaluated on SOTS-Indoor~\cite{li2018benchmarking}.

\vspace{1mm}
\noindent\textbf{Effectiveness of Individual Components.}
We construct ablations in Tab.~\ref{tab:ablation} to demonstrate the effectiveness of each component in \ours.
When we obtained $\mathrm{z}$ directly using the hazy image as the ARE’s input (``\textit{w/o} DM''), the PSNR decreases by 2.77 dB. 
This drop indicates that DM can generate a valuable amplitude spectrum to enhance dehazing performance.  
When we replace the ARE with a simple ResBlock (``\textit{w/o} ARE''), the PSNR drops by 0.31 dB. This result confirms that ARE effectively captures an amplitude residual that is essential for supervising the DM training.
When we directly remove the PCM (``\textit{w/o} PCM''), the PSNR decreases by 0.29 dB. It proves that the PCM can effectively refine the phase spectrum degradation caused by haze. 
We also explore the impact of introducing the amplitude residuals into a single scale (``\textit{w/o} MS'') and training only the DM in the second stage (``\textit{w/o} JT''). The results show that multi-scale learning and joint training can effectively enhance the model's capabilities. When all components are included, the model performance reaches 27.43 dB. 
We further explore the visual differences in~\cref{fig:case_ablation}. Without DM cannot eliminate most of the thick haze. The absence of ARE and PCM is insufficient to remove localized haze and produce clearer textures.
These demonstrate the important role of the individual components in our \ours.

\vspace{1mm}
\noindent\textbf{Effectiveness of ARE.}
To demonstrate the reliability of ARE in~\cref{fig:overview}(b), we compare 1) using vanilla self-attention~\cite{dosovitskiy2020image} and 2) using the ARE directly in RGB space to obtain amplitude residuals.
\cref{tab:ae} shows that ARE can provide more efficient and robust supervision for DM in feature space even without introducing additional parameters. 
In addition, we study the importance of mean (``\textit{w/} mean'') and standard deviation (``\textit{w/} std'') for aligning distributions in~\cref{eq:daf}. Results demonstrate that ARE (``\textit{w/} mean+std'') can explicitly align the amplitude distribution while preserving its identity.

\begin{figure}[t]
  \begin{minipage}[h]{0.26\textwidth}\small
    \centering
        \setlength{\tabcolsep}{0.6mm}{
        \centering
        \begin{tabular}{l|cc}
        \toprule
        Methods    & PSNR  & SSIM   \\
        \midrule
        \myrowcolour%
        \textit{w/o} DM &  24.66 & 0.922  \\
        Memory Bank~\cite{wu2018unsupervised} &  27.14  &  0.951 \\
        \myrowcolour%
        Sparse Coding~\cite{fan2020neural} &  25.26  &  0.931 \\
        Vanilla VQ~\cite{van2017neural} &  26.89  & 0.950 \\
        \myrowcolour%
        Diffusion Model &   \textbf{27.43} &  \textbf{0.957} \\
        \bottomrule
        \end{tabular}
        \vspace{-2mm}
        \captionof{table}{Results of different reconstruct amplitude residual methods.}
        \label{tab:diffmodel}
        }
  \end{minipage}%
  \hfill
  \begin{minipage}[h]{0.21\textwidth}\small
    \centering
    \setlength{\tabcolsep}{0.5mm}{
    \begin{tabular}{c|ccc}
    \toprule
    $T$    & PSNR  & SSIM & FLOPs(G) \\
    \midrule
    \myrowcolour%
    1 & 24.76   &  0.925 & 22.94\\
    4 & 25.51   & 0.935 & 36.49\\
    \myrowcolour%
    8 & 27.43   & 0.957 & 54.56\\
    16& 27.74  & 0.958 & 90.70 \\
    \myrowcolour%
    32 &  27.81  & 0.959 & 162.98 \\
    \bottomrule
    \end{tabular}
    \vspace{-2.3mm}
    \captionof{table}{Ablation study of the iteration steps $T$.}
    \label{tab:T}
    }
  \end{minipage}
  \vspace{-2mm}
\end{figure}

\vspace{1mm}
\noindent\textbf{Effectiveness of PCM.}
\label{pcm}
To investigate the effect of PCM in~\cref{fig:overview}(c), we compared other methods~\cite{vaswani2017attention,jaderberg2015spatial,hu2018squeeze,woo2018cbam} to refine the phase spectrum. \cref{tab:pcm} shows that our PCM has higher performance compared to the four typical attention mechanisms with a much simpler structure than others.

\vspace{1mm}
\noindent\textbf{Effectiveness of DM.}
\label{Iterations}
To demonstrate the necessity of DM, we compare it with other methods~\cite{wu2018unsupervised,fan2020neural,van2017neural} in \cref{tab:diffmodel}. Results show that the DM performs favorably, proving the advantages of reconstructing amplitude residuals.
In addition, \cref{tab:T} shows that the performance is positively correlated with the iterations steps $T$ and the gain gradually decreases when $T$ is over 8. We set $T$ to 8 after a trade-off between performance and FLOPs.
It demonstrates the effectiveness of DM and only requires fewer iterations.

\section{Conclusion}
In this paper, we introduce a new perspective of frequency-domain amplitude reconstruction to handle hazy images and propose a novel DM for unpaired image dehazing (\ours). In \ours, we recover dehazed images by performing a DM to reconstruct the amplitude spectrum consistent with the distribution of clear images.
To achieve this, we introduce an ARE used during training to provide supervision for DM training and a PCM to refine the phase spectrum. 
Such a design fully exploits the beneficial knowledge in unpaired clear images.
Experimental results show FrDiff performs favourably compared to other methods.
We hope our work can provide inspiration for other unpaired restoration tasks.

\clearpage
\clearpage
\input{supp}
\clearpage
\clearpage
{
    \small
    \bibliographystyle{ieeenat_fullname}
    \bibliography{main}
}

\end{document}

%% file: supp.tex
\clearpage
\setcounter{page}{1}
\setcounter{section}{0}
\setcounter{equation}{0}
\setcounter{figure}{0}
\setcounter{table}{0}
\setcounter{algorithm}{0}
\renewcommand{\thesection}{A\arabic{section}}
\renewcommand{\thefigure}{A\arabic{figure}}
\renewcommand{\thetable}{A\arabic{table}}
\renewcommand{\theequation}{A\arabic{equation}}
\renewcommand{\thealgorithm}{A\arabic{algorithm}}
\maketitlesupplementary

In this supplementary material,~\cref{A} provides the motivation analysis of the reconstructed amplitude in the frequency domain.
\cref{B} illustrates the detailed architecture of the dehazing network and analyzes its efficiency. 
\cref{C} describes the detailed training and inference algorithms.
\cref{AA} highlights the advantages of using the unpaired data training paradigm in this work.
\cref{D} and \cref{E} describe the dataset used and the training details, respectively.
\cref{H} analyzes the limitations of our method. 
Finally,~\cref{I} shows more quantitative and qualitative comparison results.

\section{Motivation Analysis}
\label{A}

As described in~\cref{sec:intro} of the main paper, the illumination contrast is manifested by the amplitude spectrum, while the texture structure information is manifested by the phase spectrum. Haze mainly affects the illumination contrast, while the structural information is immune to haze degradation. That is to say, that haze degradation is transferred with the exchange of the amplitude spectrum. Therefore, the key motivation of our FrDiff is to make the diffusion model (DM) reconstruct the amplitude of haze-free images in the unpaired training setting. 

To justify this motivation, we follow the assumption of the classical dark channel prior~\cite{he2010single}, and analyze it by studying the statistical properties of the dark channel before and after the exchange of the amplitude spectrum. Specifically, as shown in Fig.~\ref{fig:teaser}(a) of the main paper, we use the FFT and IFFT operations to replace the amplitude spectrum of the haze images (\ie,~Hazy) with the amplitude spectrum of the clear images (\ie,~Clear) on the SOTS-Indoor~\cite{li2018benchmarking} dataset. By doing so, we can obtain a set of synthetic clear images (\ie,~SynClear). Then, we follow~\cite{li2018benchmarking} using the patch with the size of $15\times 15$ to calculate their dark channels.

\begin{figure}[t]
\begin{center}
\includegraphics[width=1.0\linewidth,page=1]{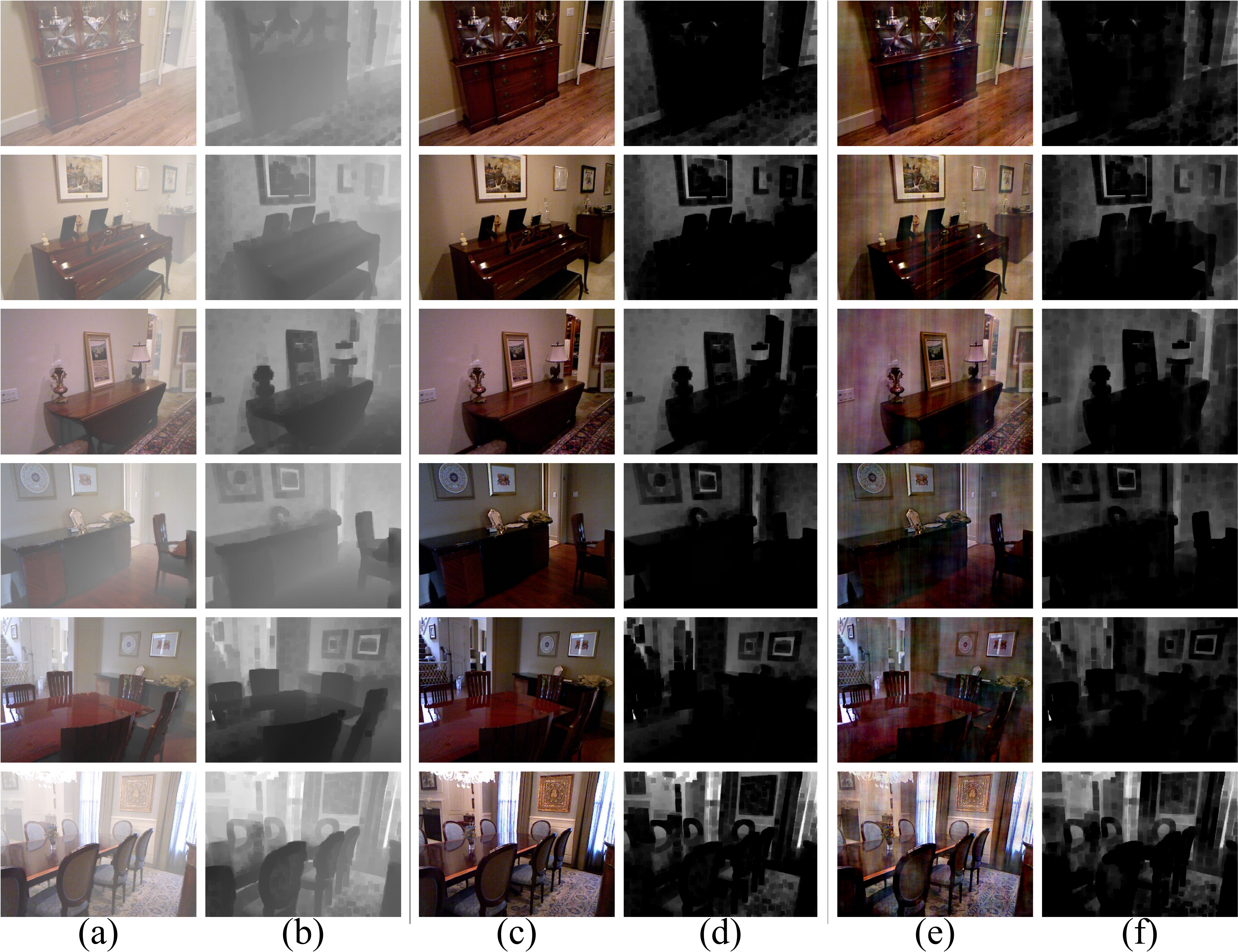}
\end{center}
\vspace{-6mm}
\caption{Example images in SOTS-Indoor~\cite{li2018benchmarking} dataset. (a), (c), and (e) are the haze images (\ie,~Hazy), the clear images (\ie,~Clear), and the synthetic clear images (\ie,~SynClear), respectively. (b), (d), and (f) are their corresponding dark channels. The dark channel of the SynClear image is more similar to the Clear image indicating that the replacement of the amplitude spectrum is effective in removing the haze.}
\label{fig:supp-motivation}
\end{figure}

\cref{fig:supp-motivation} shows several example images and the corresponding dark channels. Figs.~\ref{fig:supp-motivation}(a), (c), and (e) are the Hazy images, the Clear images, and the SynClear images, respectively. Figs.~\ref{fig:supp-motivation}(b), (d), and (f) are their corresponding dark channels. Visually, the intensity of the dark channel is a rough approximation of the thickness of the haze. The dark channel of a Hazy image will have higher intensity in regions with denser haze (see~\cref{fig:supp-motivation}(b)).
As the amplitude spectrum of the Hazy image is replaced with the amplitude spectrum of the Clear image, the dark channels of the SynClear image are closer to the dark channels of the Clear image (see Figs.~\ref{fig:supp-motivation}(d) and (f)). This proves that the data properties of the synthetic clear image are closer to those of the clear image.

\begin{figure*}[th]
\begin{center}
\includegraphics[width=1.0\linewidth,page=2]{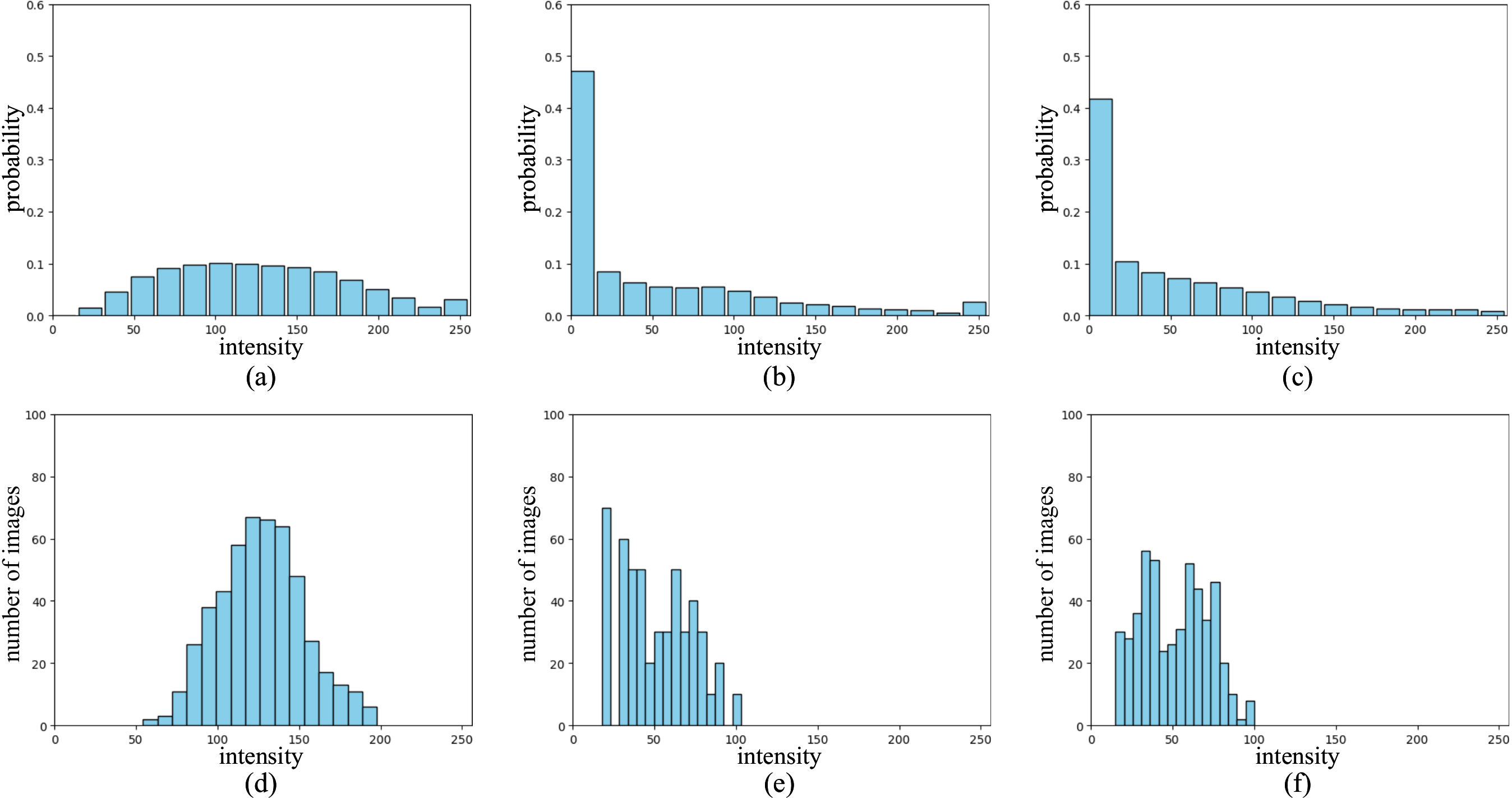}
\end{center}
\vspace{-6mm}
\caption{Statistics of the dark channels. (a), (b), and (c) are the histogram of the pixel intensities in all of the dark channels of the haze images (\ie,~Hazy), the clear images (\ie,~Clear), and the synthetic clear images (\ie,~SynClear) in the dataset, respectively. (d), (e), and (f) are the corresponding histogram of the average intensity of each dark channel (each bin stands for 16 intensity levels). The dark channel statistics indicate that the synthetic clear image obtained by replacing the amplitude spectrum is closer to the clear image distribution.}
\label{fig:supp-motivation2}
\end{figure*}

In addition, Figs.~\ref{fig:supp-motivation2}(a), (b), and (c) are the histogram of the pixel intensities in all of the dark channels of the haze images (\ie,~Hazy), the clear images (\ie,~Clear), and the synthetic clear images (\ie,~SynClear) in the dataset, respectively. Figs.~\ref{fig:supp-motivation2}(d), (e), and (f) are the corresponding histograms of the average intensity of each dark channel (each bin stands for 16 intensity levels). We have the following observations:
\begin{itemize}
    \item 1) The intensity of the pixels in the dark channel of the Hazy images mostly falls between 50-200 (see Fig.~\ref{fig:supp-motivation2}(a)). In contrast, the Clear and SynClear images have mostly zero values and about 75 percent of the pixels are below 25 (see Figs.~\ref{fig:supp-motivation2}(b) and (c)).
    \item 2) The average intensity of each dark channel of the Hazy images also mostly falls between 100-150 (see Fig.~\ref{fig:supp-motivation2}(d)), while the average intensity of each dark channel for both Clear and SynClear images is below 100 (see Figs.~\ref{fig:supp-motivation2}(e) and (f)).
\end{itemize}
In general, this statistic not only provides very strong support for the dark channel prior~\cite{he2010single}, but also demonstrates that the illumination contrast (\eg,~Haze) is manifested by the amplitude spectrum, and replacing the amplitude spectrum from the clear image can effectively remove the haze.

Therefore, this unique frequency domain property provides inspiration for haze removal. Our key insight of FrDiff, \ie, allowing DM to learn the amplitude of haze-free images in the unpaired training setting, is justified and has great potential.

\section{Architecture Details}
\label{B}

As described in Sec.~\ref{FrDiff} of the main paper. To ensure the non-local feature capture capability and the reconstruction capability of the dehazing process, we adopt NAFNet~\cite{chen2022simple}, a simple UNet-based architecture, as our dehazing network.

\begin{figure}[t]
\begin{center}
\includegraphics[width=1.0\linewidth,page=3]{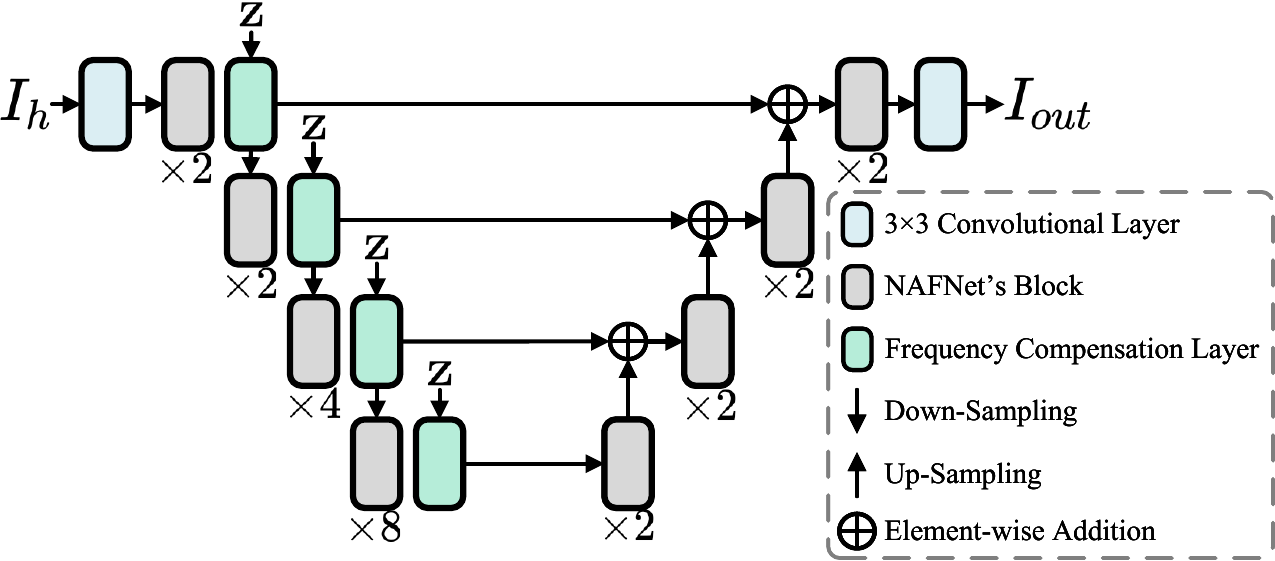}
\end{center}
\vspace{-6mm}
\caption{Network structure of dehazing network.}
\label{fig:supp-net}
\end{figure}

Specifically, we illustrate the detailed architecture of the lightweight U-Net as shown in Fig.~\ref{fig:supp-net}. The entire network is based on the U-Net architecture. We follow the existing approach~\cite{chen2022simple} to extract features by stacking some NAFNet`s Blocks on each scale, where the number of basic blocks is marked. In each scale, we also incorporate the Frequency Compensation Layer (see~\cref{fig:overview}(a) of the main paper) in the contracting path (up-to-down) of UNet and input amplitude residual $\mathrm{z}$ into these layers. The dehazed image is outputted through the expanding path (down to up) of UNet. The parameters of the dehazing network are 8.69M. 

In addition, we use a neural network consisting of 5 stacked ResBlocks, denoted as $\epsilon_\theta$, to estimate the noise. The purpose of using ResBlocks as the denoising network is to ensure the same resolution of inputs and outputs while minimizing the model parameters. The parameters of the denoising network are 0.07M and the FLOPs required per iteration is about 4.5G. 

On 3090 GPU with $256\times256$ input, the runtime is 0.04s. Future we’ll further accelerate the denoising process by one-step distillation.

\section{Algorithm}
\label{C}

The first and second stage training algorithms for FrDiff are shown in Alg.~\ref{alg:train_S1} and Alg.~\ref{alg:train_S2}, respectively. The inference algorithm for FrDiff is shown in Alg.~\ref{alg:inference}.

\section{Advantages of Unpaired Training}
\label{AA}
The field of unsupervised training for image dehazing can be divided into pseudo-label-based approaches~\cite{chen2021psd,wu2023ridcp,fang2025real,shao2020domain,chen2024prompt} and unpaired training based approaches~\cite{qiao2023learning,liu2020end,engin2018cycle,yang2022self,park2020contrastive,chen2022unpaired,wang2024ucl,wang2024odcr}. 

The pseudo-label-based approaches~\cite{chen2021psd,wu2023ridcp,fang2025real,shao2020domain,chen2024prompt} focus on mining and modeling the characteristics of real haze, and usually improves its performance in real-world haze scenarios by introducing an additional re-hazing pipeline. However, the design of a reasonable pipeline critically hinges on domain expertise. Without expert insights, the performance of such methods may falter due to the inherent challenges in capturing domain-specific characteristics. In contrast, unpaired training-based approaches~\cite{qiao2023learning,liu2020end,engin2018cycle,yang2022self,park2020contrastive,chen2022unpaired,wang2024ucl,wang2024odcr} aims to directly learn the mapping of the haze domain to the clear domain through tailored training strategies, and thus more concise and promising.

\begin{algorithm}[t]
	\caption{ FrDiff Training: Stage One}
	\label{alg:train_S1}
	\textbf{Input}: ARE, dehazing network. \\
	\textbf{Output}: Trained dehazing network. \\
        \vspace{-4mm}
	\begin{algorithmic}[1] 
            \FOR{$I_{h}$,  $I_{c}$ }
            \STATE    $\mathrm{z}=\operatorname{ARE}(I_{h},I_{c}). $ (paper Eqs.~(\textbf{\textcolor{blue}{1}})-(\textbf{\textcolor{blue}{3}}))
            \STATE $I_{out} = \operatorname{DehazingNetwork}(I_{h},\mathrm{z})$
            \STATE Calculate $\mathcal{L}_{s1}$ loss (paper Eq.~(\textbf{\textcolor{blue}{6}})).
            \ENDFOR     
		\STATE Output the trained dehazing network.
	\end{algorithmic}
\end{algorithm}

\begin{algorithm}[t]
	\caption{ FrDiff Training: Stage Two}
	\label{alg:train_S2}
	\textbf{Input}: ARE, trained dehazing network,, denoising network, $\beta_t (t\in[1,T])$. \\
	\textbf{Output}: Trained denoising network, trained dehazing network. \\
        \vspace{-4mm}
	\begin{algorithmic}[1] 
		\STATE Init: $\alpha_t=1-\beta_t$, $\bar{\alpha}_T=\prod_{i=0}^T \alpha_i$.
            \STATE Init: The dehazing network copies the parameters of trained dehazing network. 
            \FOR{$I_{h}$,  $I_{c}$ }
            \STATE    $\mathrm{z}=\operatorname{ARE}(I_{h},I_{c}). $ (paper Eqs.~(\textbf{\textcolor{blue}{1}})-(\textbf{\textcolor{blue}{3}}))
            \STATE \textbf{Diffusion Process}:
            \STATE  We sample $\mathrm{z}_{T}$ by $q\left(\mathrm{z}_T \mid               \mathrm{z}\right)=\mathcal{N}\left(\mathrm{z}_T; \sqrt{\bar{\alpha}_T} \mathrm{z},\left(1-\bar{\alpha}_T\right) \mathbf{I}\right)$  (paper Eq.~(\textbf{\textcolor{blue}{7}})) 
            \STATE \textbf{Denoising Process}:
            \STATE $\hat{\mathrm{z}}_T = \mathrm{z}_{T}$
            \STATE $A_h=\operatorname{FFT}(\operatorname{Conv-Block}(I_{h}))$
            \FOR{$t=T$ to $1$ }
            \STATE $\hat{\mathrm{z}}_{t-1}\!=\!\frac{1}{\sqrt{\alpha_t}}(\hat{\mathrm{z}}_t-\frac{1-\alpha_t}{\sqrt{1-\bar{\alpha}_t}} \epsilon_\theta(\hat{\mathrm{z}}_t, A_h, t))\!+\!\sqrt{1\!-\!\alpha_t} \epsilon_t$
            (paper Eq.~(\textbf{\textcolor{blue}{10}}))            
            \ENDFOR
            \STATE $\hat{\mathrm{z}}=\hat{\mathrm{z}}_{0}$
            \STATE $I_{out} = \operatorname{DehazingNetwork}(I_{h},\hat{\mathrm{z}})$
            \STATE Calculate $\mathcal{L}_{s2}$ loss (paper Eq.~(\textbf{\textcolor{blue}{11}})).
            \ENDFOR     
		\STATE Output the trained denoising network and trained dehazing network.
	\end{algorithmic}
\end{algorithm}

\begin{algorithm}[t]
	\caption{ FrDiff Inference}
	\label{alg:inference}
	\textbf{Input}: Trained denoising network, trained dehazing network, $\beta_t (t\in[1,T])$, hazy images $I_{h}$. \\
	\textbf{Output}: Dehazed images $I_{out}$. \\
        \vspace{-4mm}
	\begin{algorithmic}[1] 
		\STATE Init: $\alpha_t=1-\beta_t$, $\bar{\alpha}_T=\prod_{i=0}^T \alpha_i$.
            \STATE \textbf{Denoising Process}:
            \STATE Sample $\mathrm{z}_T \sim \mathcal{N}(0,1)$
            \STATE $\hat{\mathrm{z}}_T = \mathrm{z}_{T}$
            \STATE $A_h=\operatorname{FFT}(\operatorname{Conv-Block}(I_{h}))$
            \FOR{$t=T$ to $1$ }
            \STATE  $\hat{\mathrm{z}}_{t-1}\!=\!\frac{1}{\sqrt{\alpha_t}}(\hat{\mathrm{z}}_t-\frac{1-\alpha_t}{\sqrt{1-\bar{\alpha}_t}} \epsilon_\theta(\hat{\mathrm{z}}_t, A_h, t))\!+\!\sqrt{1\!-\!\alpha_t} \epsilon_t$
            (paper Eq.~(\textbf{\textcolor{blue}{10}}))                 
            \ENDFOR
            \STATE $\hat{\mathrm{z}}=\hat{\mathrm{z}}_{0}$
            \STATE $I_{out} = \operatorname{DehazingNetwork}(I_{h},\hat{\mathrm{z}})$
   
		\STATE Output dehazed images $I_{out}$.
	\end{algorithmic}
\end{algorithm}

\section{More Dataset Details}
\label{D}
We evaluate our method on widely-used RESIDE~\cite{li2018benchmarking}, I-HAZE~\cite{ancuti2018haze}, and Fattal's~\cite{fattal2014dehazing} datasets, which cover synthetic, artificial, and real-world images. 
Specifically, the RESIDE~\cite{li2018benchmarking} dataset contains several subsets: (a) ITS, which contains 13,990 synthetic indoor hazy-clear pairs. (b) SOTS-Indoor and SOTS-Outdoor, which contain 500 synthetic indoor/outdoor hazy-clear pairs. (c) HSTS-Synth and HSTS-Real, which contain 10 synthetic hazy-clear pairs and 10 real-world hazy image without ground-truth images, respectively. (d) URHI, which contains over 4,000 real hazy image without ground-truth images. I-HAZE~\cite{ancuti2018haze} dataset contains 35 artificial hazy-clear pairs produced by professional haze generators. Fattal's~\cite{fattal2014dehazing} dataset includes 31 real-world hazy images in various scenes. 

For fair comparisons, we follow the previous works~\cite{yang2022self,wang2024odcr} to use ITS dataset from RESIDE as the training set. We validate the performance of FrDiff on synthetic data using the SOTS-Indoor, SOTS-Outdoor, and HSTS-Synth datasets with ground truth. We validate the performance of FrDiff on real-world data using the HSTS-Real, Fattal's, and URHI datasets without ground truth.

\section{More Training Details}
\label{E}
During training, we follow existing works~\cite{chen2024hierarchical,xia2023diffir} to adopt a two-stage strategy to optimize our model. 

As shown in~\cref{fig:training}(a) of the main paper, in stage one, we utilize the proposed amplitude residual encoder (ARE) to obtain amplitude residual $\mathrm{z}$, and train the dehazing network for haze removal. 
In this case, the $\mathrm{z}$ is directly fed into the dehazing network without involving the diffusion and denoising processes.
It is infeasible to supervise model training with the strict pixel-level loss function under unpaired data setting. Therefore,  we follow existing works~\cite{park2020contrastive,wang2024odcr,wang2024ucl}, using the same adversarial loss $\mathcal{L}_{GAN}$ and patchwise contrast loss $\mathcal{L}_{PatchNCE}$ to encourage the dehazed results have the same domain distribution as the clear images, and the hyper-parameters $\lambda_{GAN}$ and $\lambda_{PatchNCE}$ are all set to 1.

As shown in~\cref{fig:training}(b) of the main paper, in stage two, we joint train diffusion model and dehazing network from stage one to reconstruct the amplitude residual for enhancing the dehazing process. In this stage, the $\mathrm{z}$ first adds noise to output $\mathrm{z}_T$ through the diffusion process, and then reconstructs the amplitude residual $\hat{\mathrm{z}}$ through multiple denoising processes. 
We additionally include the diffusion loss $\mathcal{L}_{diff}$ based on the $\mathcal{L}_{GAN}$ and $\mathcal{L}_{PatchNCE}$, and the hyper-parameter $\lambda_{diff}$ is set to 1. 

In both training stages, we use Adam~\cite{kingma2014adam} optimizer with $\beta_{1}=0.9$ and $\beta_{2}=0.999$, learning rate is $1\times 10^{-4}$. The epoch number is 200. The batch size is $8$ and the input patch size is $256\times 256$ and augments the data with random horizontal and vertical flips. The input size of $256\times256$ is randomly cropped from all training images in an unpaired learning procedure.

\section{Limitation}
\label{H}
Although our FrDiff can reconstruct frequency domain features well for handling haze, when the resolution of the input image is larger, the resolution of the amplitude residuals reconstructed by the diffusion model also increases. This means that the computational effort of the diffusion model will increase. Therefore, it is expected to make the diffusion model learn a set with a fixed number of amplitude spectral priors to reconstruct the frequency domain features so as to avoid increasing computational costs significantly. 

In addition, removing the haze by reconstructing the spectrum ignores the local spatial differences  of the haze, and the future promises to further enable the model to handle spatially varying haze of different thicknesses.

\section{More Results}
\label{I}
In this section, we first provide more quantitative results and spectral visualisations to validate the effectiveness of FrDiff. Then, we analyze the sensitivity of the hyper-parameters in the loss function. Finally, we show more visualization results.

\vspace{-2mm}
\paragraph{Results on O-HAZE~\cite{ancuti2018haze2} dataset.}
To further demonstrate the generalization ability of our FrDiff, we add comparisons on O-HAZE~\cite{ancuti2018haze2} in~\cref{tab:OHAZE}.
Experimental results show that FrDiff achieves higher performance compared to prevailing unpaired data-based training methods.

\begin{table}[h]
\centering
\setlength{\tabcolsep}{0.8mm}{
  \begin{tabular}{ l  | c  c  c  c}
    \toprule
    Methods       &  \cellcolor{LightGray}$D^4$\cite{yang2022self}  &  UCL-Dehaze\cite{wang2024ucl}  & \cellcolor{LightGray}ODCR\cite{wang2024odcr}  & FrDiff \\
    \midrule
    PSNR       & \cellcolor{LightGray}16.92   &  15.57  &  \cellcolor{LightGray}17.46   & \textcolor{red}{18.22}\\
    SSIM       &  \cellcolor{LightGray}0.607  &  0.566  &   \cellcolor{LightGray}0.632 & \textcolor{red}{0.648}\\ 
    FADE       & \cellcolor{LightGray}1.358   &  1.715  &  \cellcolor{LightGray}0.970   & \textcolor{red}{0.744}\\
    MUSIQ       &  \cellcolor{LightGray}63.042  &  62.432  &   \cellcolor{LightGray}63.388 & \textcolor{red}{63.981}\\
    \bottomrule
  \end{tabular}}
  \caption{Quantitative comparison with SOTA methods on O-HAZE~\cite{ancuti2018haze2} dataset.}
\label{tab:OHAZE}
\end{table}

\begin{figure}[h]
  \centering
    \includegraphics[width=1.0\linewidth,page=1]{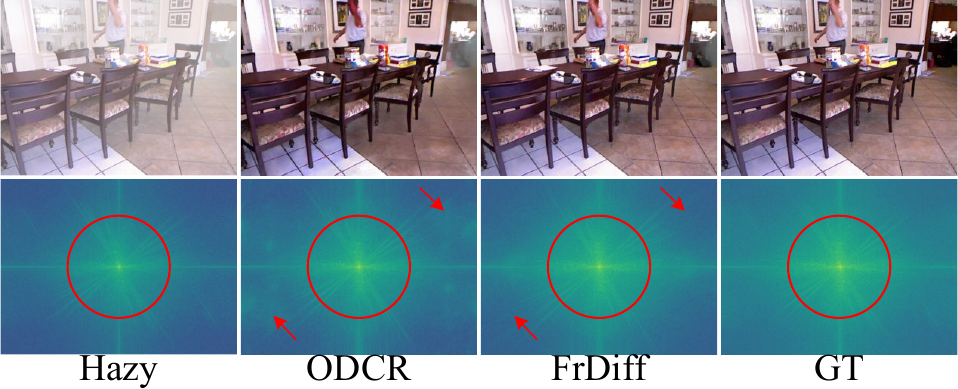}
    \vspace{-3mm}
   \caption{Visual results on SOTS-Indoor~\cite{li2018benchmarking} and SOTS-Outdoor~\cite{li2018benchmarking} datasets. 
   Zoom in to see better visualization.}
   \label{fig:fre}
\end{figure}

\vspace{-2mm}
\paragraph{Spectral Visualizations}
To prove the reliability of FrDiff, we present the spectral visualizations in~\cref{fig:fre}. FrDiff can effectively reconstruct low-frequency illuminance (red circle) and eliminate the high-frequency spectral bias (red arrow). In addition, we measure the L2 similarity($\downarrow$) of the frequency features with GT to show the superiority (ODCR:6.257, FrDiff:2.156).

\vspace{-2mm}
\paragraph{Effect of Hyper-parameters.}
To explore the influence of hyper-parameters used in~\cref{loss2}. We discuss the different $\lambda_{GAN}$, $\lambda_{PatchNCE}$, and $\lambda_{diff}$ as shown in~\cref{table:loss}. It is worth noting that each time we let only one hyper-parameter change and set the remaining two to 1. Proper $\lambda_{GAN}$ can provide effective supervision of model training. The impact of $\lambda_{PatchNCE}$ is insignificant since the haze-unrelated information is introduced at the same time. The performance is positively correlated with the $\lambda_{diff}$, demonstrating the ability of the denoising network to reconstruct magnitude residuals. However, a larger $\lambda_{diff}$ may lead to instability in training.
After a trade-off between performance and training stability, we choose 1 as the value of $\lambda_{GAN}$, $\lambda_{PatchNCE}$, and $\lambda_{diff}$.

\begin{table}[h]
\centering
\begin{tabular}{c |c || c | c|| c | c}
\toprule
$\lambda_{GAN}$ &  PSNR & $\lambda_{PatchNCE}$ &  PSNR & $\lambda_{diff}$ &  PSNR \\
\midrule
\myrowcolour%
0.1         & 36.39 & 0.1          & 36.43   &0.1          & 36.31  \\
1          & 36.54 & 1         & 36.54     & 1          & 36.54         \\
\myrowcolour%
10          & 36.53 & 10          & 36.50    & 10          & 36.58         \\
\bottomrule
\end{tabular}
\caption{Results of different $\lambda_{GAN}$, $\lambda_{PatchNCE}$, and $\lambda_{diff}$ on SOTS-Indoor~\cite{li2018benchmarking} dataset.}
\label{table:loss}
\end{table}

\begin{figure*}[t]
\begin{center}
\includegraphics[width=1.0\linewidth,page=4]{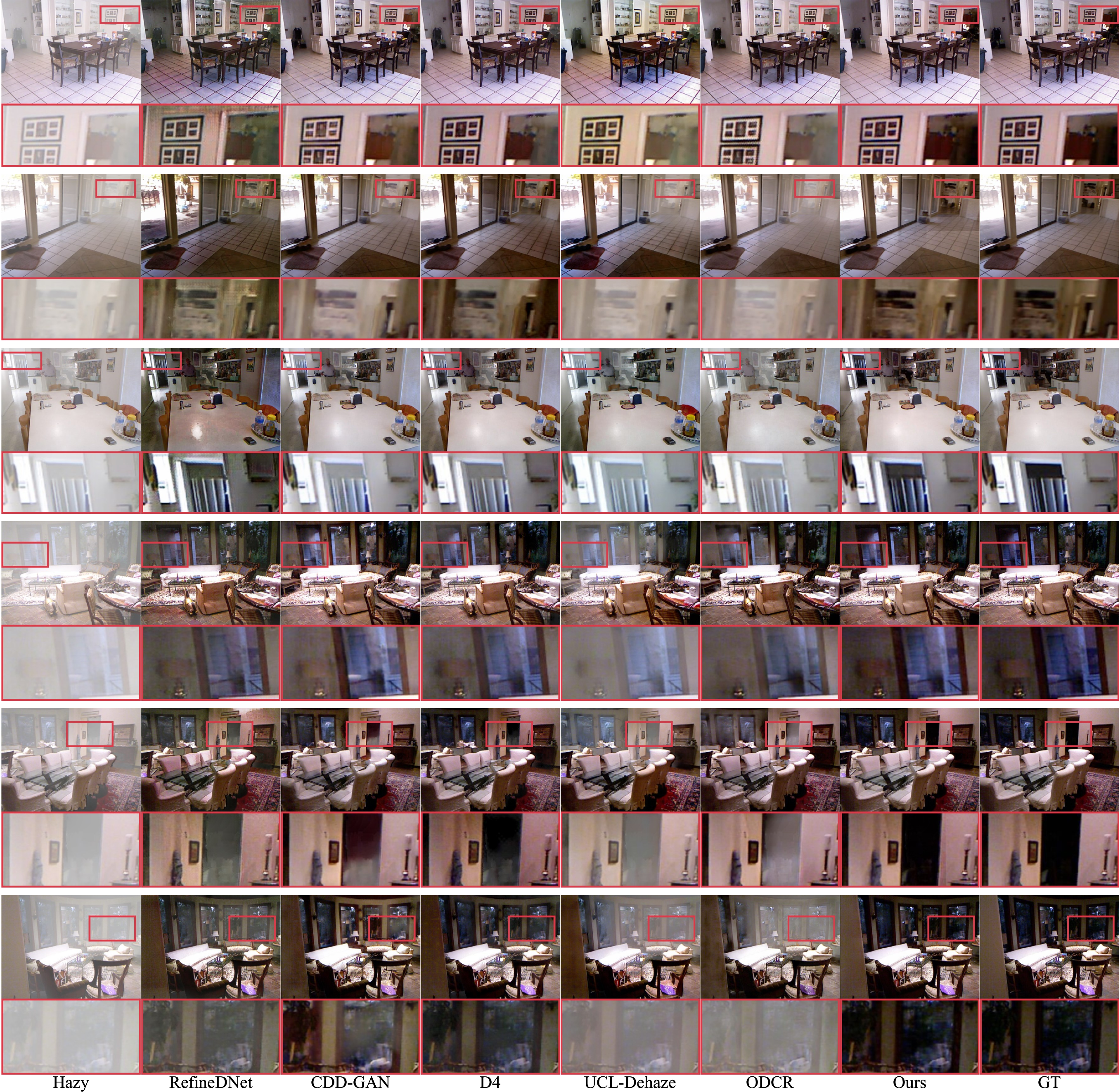}
\end{center}
\caption{Visual results on SOTS-Indoor~\cite{li2018benchmarking} dataset. The method is shown at the bottom of each case. Zoom in to see better visualization.}
\label{fig:supp-indoor1}
\end{figure*}

\begin{figure*}[t]
\begin{center}
\includegraphics[width=1.0\linewidth,page=5]{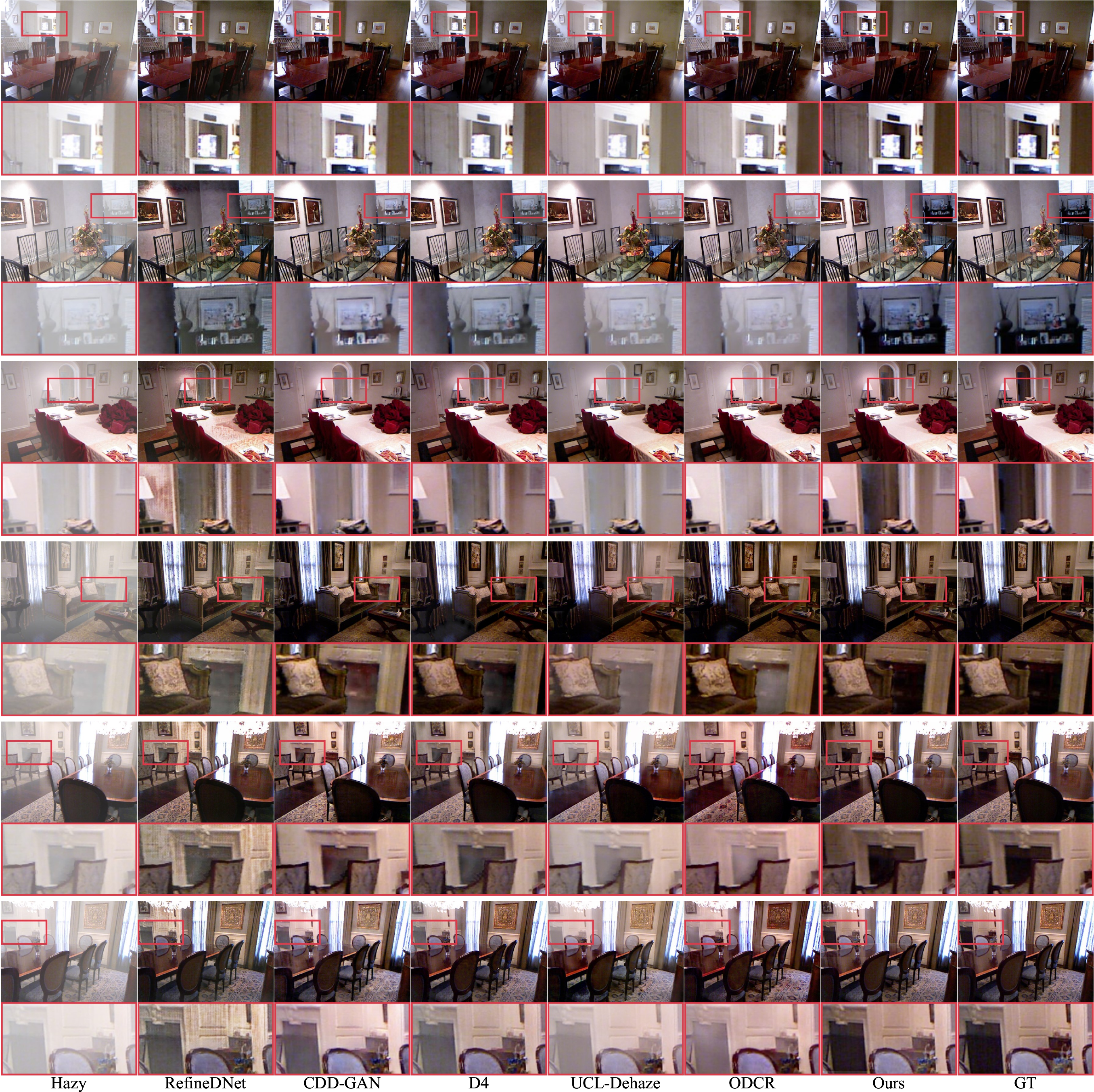}
\end{center}
\caption{Visual results on SOTS-Indoor~\cite{li2018benchmarking} dataset. The method is shown at the bottom of each case. Zoom in to see better visualization.}
\label{fig:supp-indoor2}
\end{figure*}

\begin{figure*}[t]
\begin{center}
\includegraphics[width=1.0\linewidth,page=6]{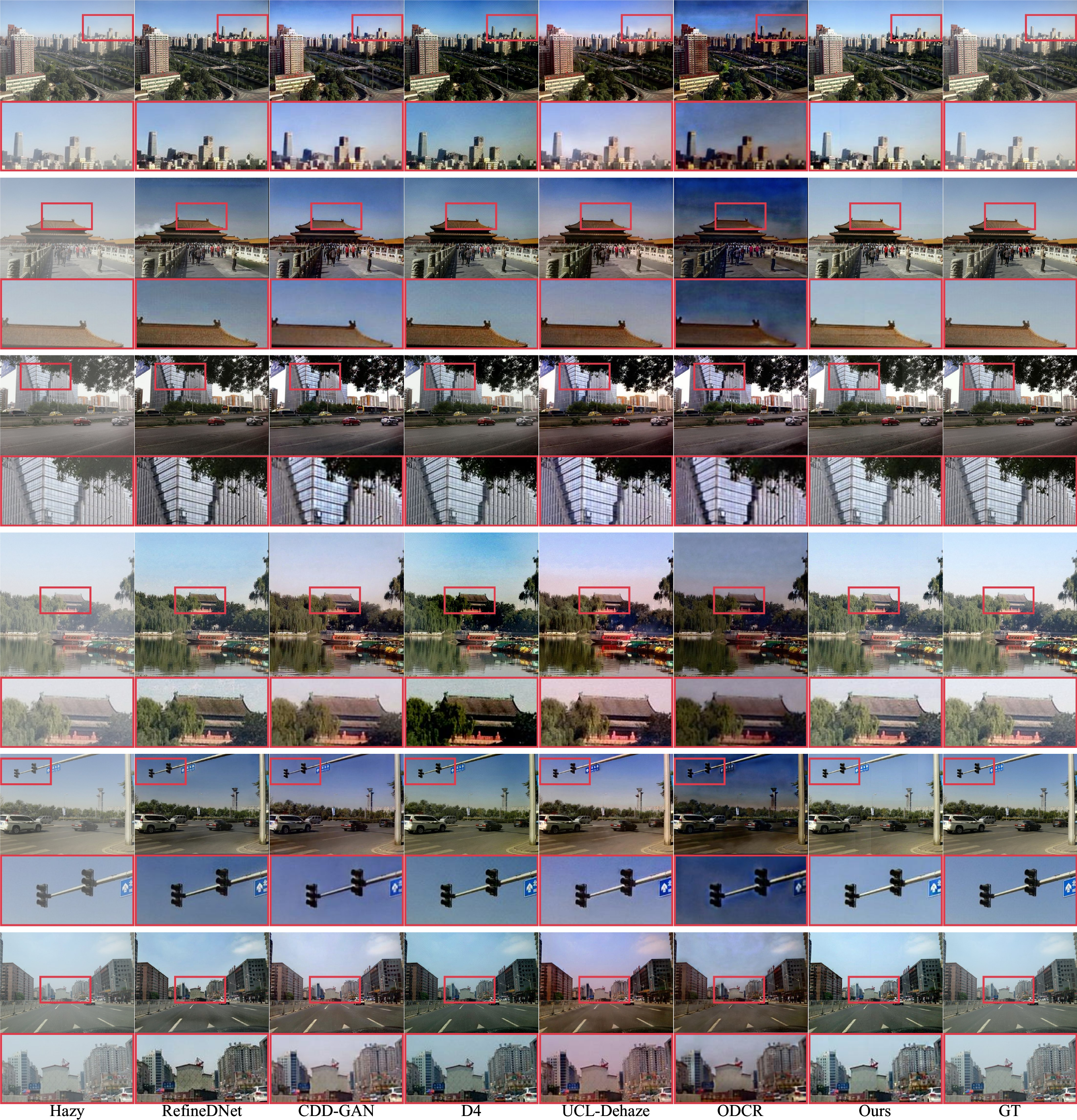}
\end{center}
\caption{Visual results on SOTS-Outdoor~\cite{li2018benchmarking} dataset. The method is shown at the bottom of each case. Zoom in to see better visualization.}
\label{fig:supp-outdoor1}
\end{figure*}

\begin{figure*}[t]
\begin{center}
\includegraphics[width=1.0\linewidth,page=7]{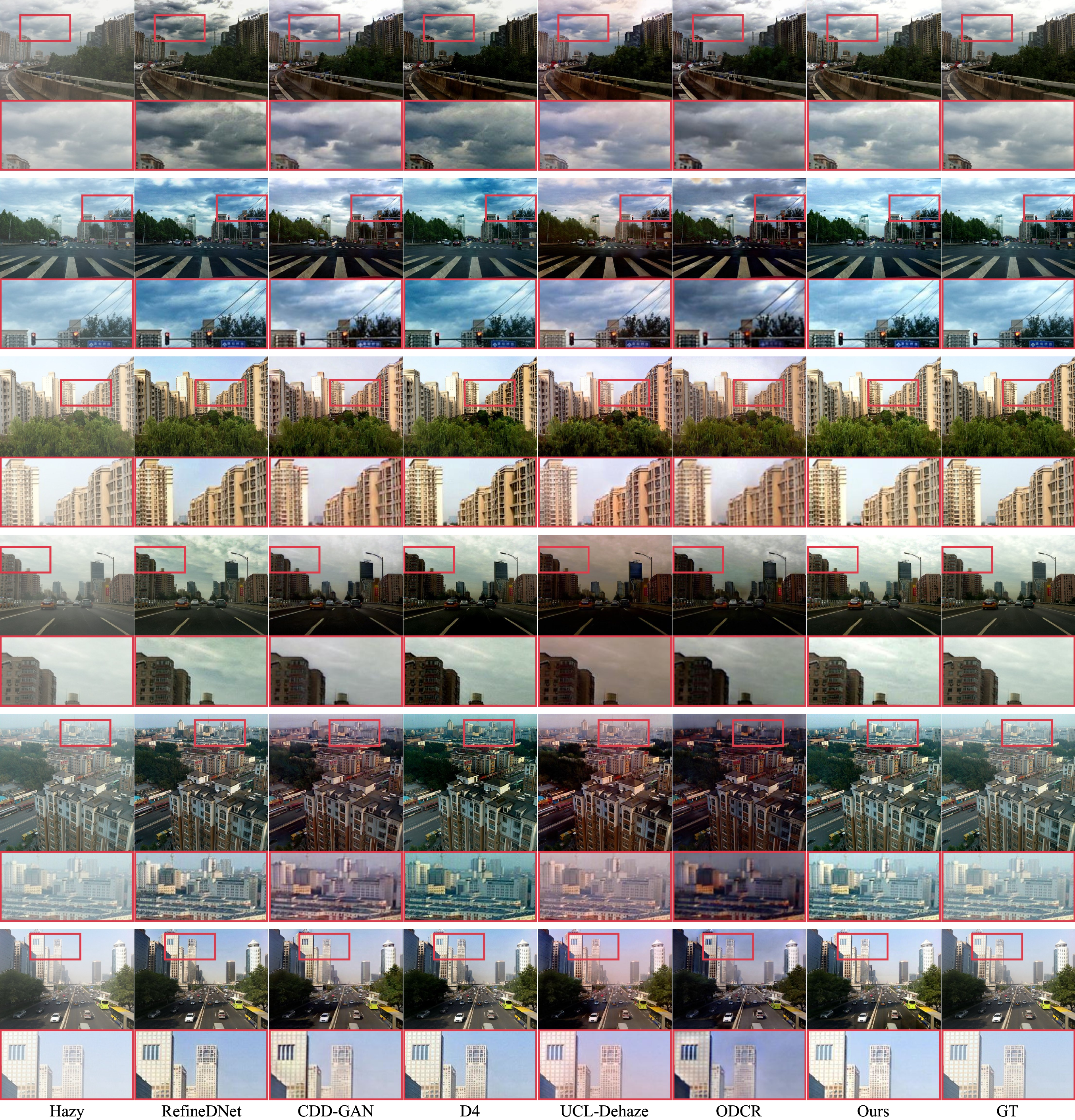}
\end{center}
\caption{Visual results on SOTS-Outdoor~\cite{li2018benchmarking} dataset. The method is shown at the bottom of each case. Zoom in to see better visualization.}
\label{fig:supp-outdoor2}
\end{figure*}

\begin{figure*}[t]
\begin{center}
\includegraphics[width=1.0\linewidth,page=8]{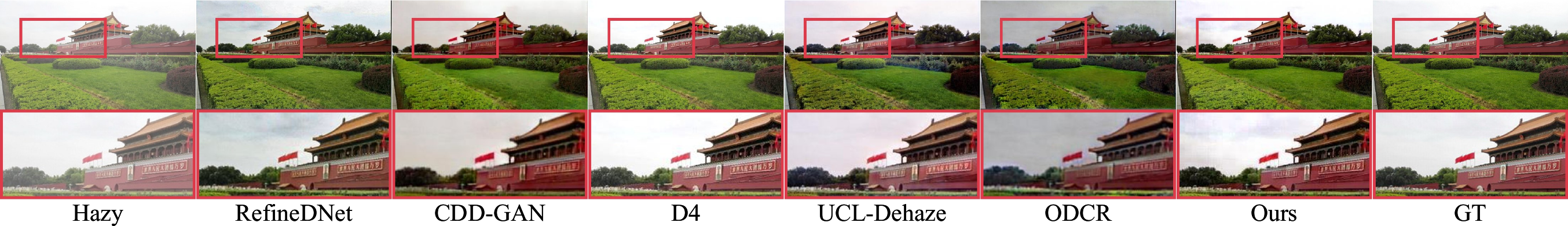}
\end{center}
\caption{Visual results on HSTS-Synth~\cite{li2018benchmarking} dataset. The method is shown at the bottom of each case. Zoom in to see better visualization.}
\label{fig:supp-HSTS}
\end{figure*}

\begin{figure*}[t]
\begin{center}
\includegraphics[width=1.0\linewidth,page=9]{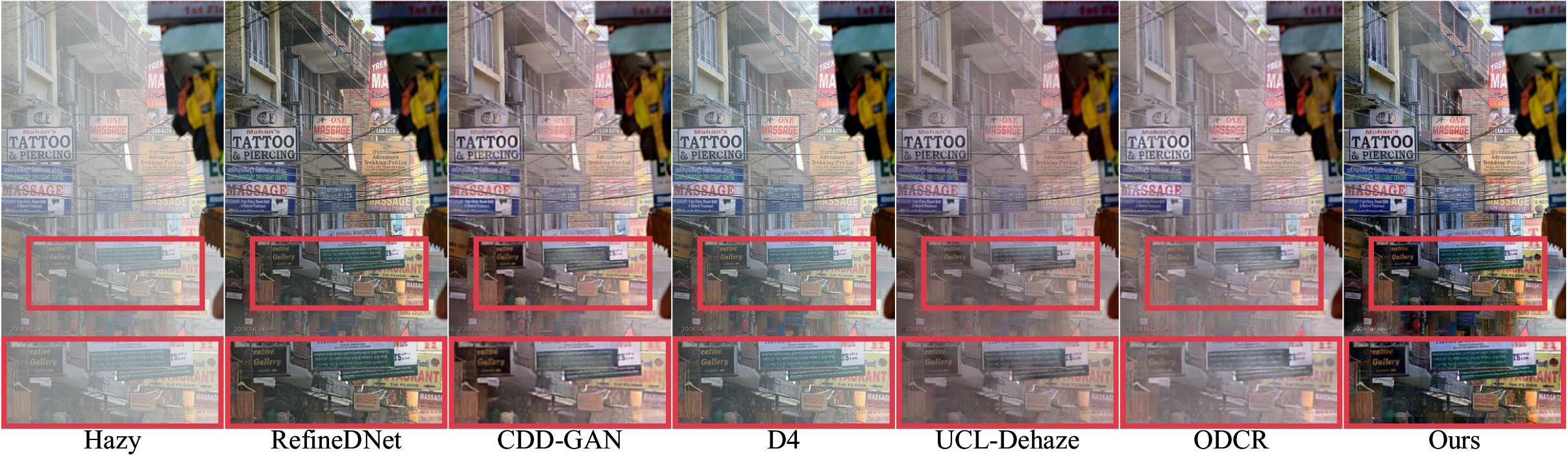}
\end{center}
\caption{Visual results on Fattle's~\cite{fattal2014dehazing} dataset. The method is shown at the bottom of each case. Zoom in to see better visualization.}
\label{fig:supp-Fattle}
\end{figure*}

\begin{figure*}[t]
\begin{center}
\includegraphics[width=1.0\linewidth,page=10]{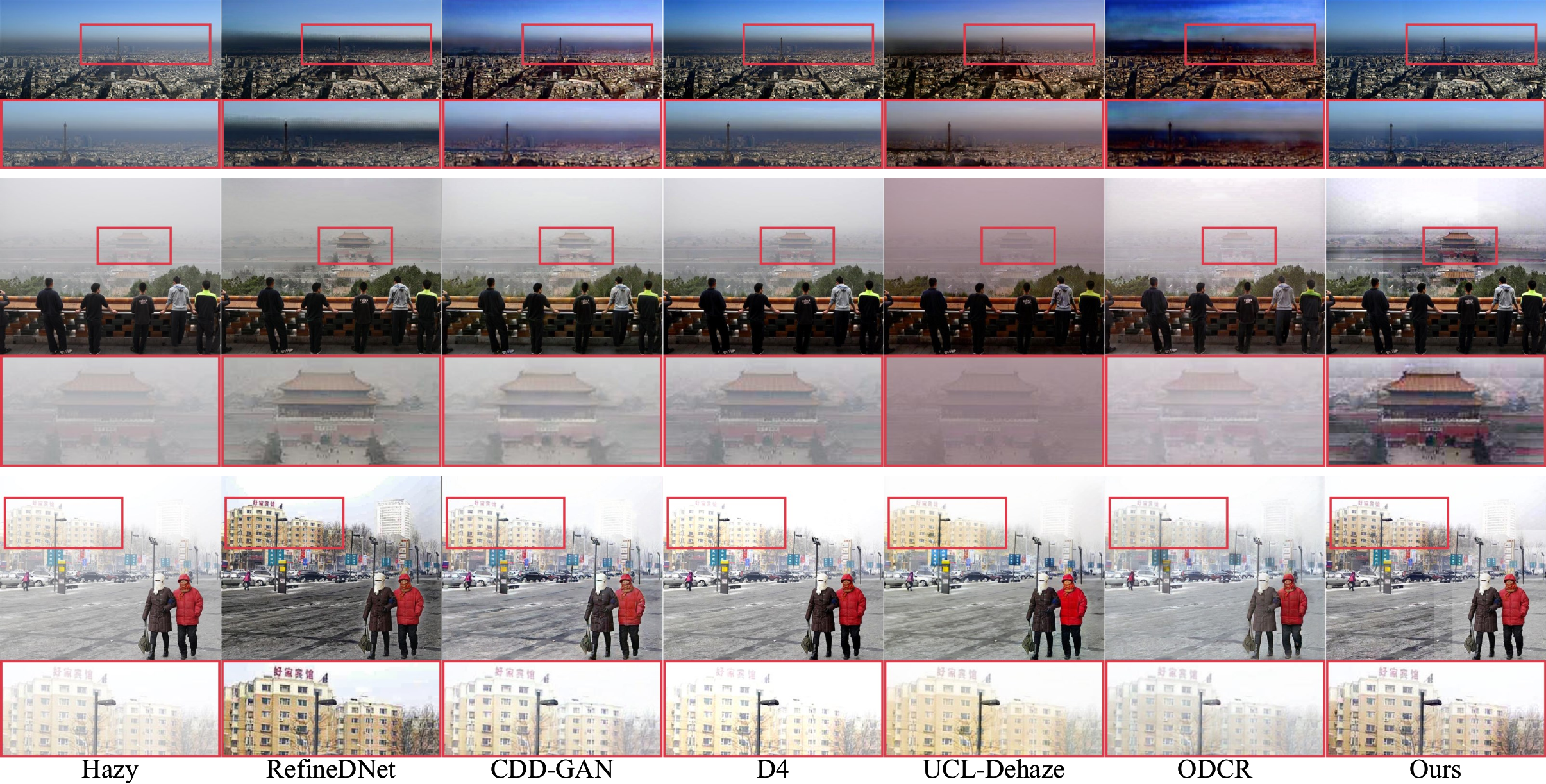}
\end{center}
\caption{Visual results on URHI~\cite{li2018benchmarking} dataset. The method is shown at the bottom of each case. Zoom in to see better visualization.}
\label{fig:supp-URHI}
\end{figure*}

\vspace{-2mm}
\paragraph{More Visual Results.}
To further verify the effectiveness of our method, we show more comparison results among the proposed FrDiff and other advanced methods on these benchmarks. The results on \textbf{SOTS-Indoor}~\cite{li2018benchmarking} dataset are shown in Figs.~\ref{fig:supp-indoor1} and~\ref{fig:supp-indoor2}. The results on \textbf{SOTS-Outdoor}~\cite{li2018benchmarking} dataset are shown in Figs.~\ref{fig:supp-outdoor1} and~\ref{fig:supp-outdoor2}. The results on \textbf{HSTS-Synth}~\cite{li2018benchmarking}, \textbf{Fattle's}~\cite{fattal2014dehazing}, and \textbf{URHI}~\cite{li2018benchmarking} dataset are shown in Fig.~\ref{fig:supp-HSTS}, Fig.~\ref{fig:supp-Fattle}, and Fig.~\ref{fig:supp-URHI}, respectively.

